\pdfoutput=1
\documentclass[times, review, 10pt]{elsarticle}

\usepackage[letterpaper, 
            top=4.3cm, 
            bottom=4.3cm, 
            left=4.8cm, 
            right=4.8cm, 
            footskip=2.6cm]{geometry} % 设置页边距和页码位置

\usepackage{setspace}
\usepackage{amsmath}
\usepackage{amssymb}
\usepackage{bm}          % 粗体数学符号
\usepackage{booktabs}    % 三线表
\usepackage{makecell}    % 单元格多行
\usepackage{adjustbox}   % 表格适配
\usepackage{multirow}    % 跨行
\usepackage{tabularx}    % 自动列宽
\usepackage{graphicx}    % 图片
\usepackage{algorithmic}
\usepackage{algorithm}
\usepackage{array}
\usepackage{subfig}
\usepackage{textcomp}
\usepackage{stfloats}
\usepackage{url}
\usepackage{verbatim}
\usepackage[shortcuts]{extdash}
\usepackage[font=footnotesize]{caption}
\biboptions{numbers,sort&compress} % 采用 Vancouver/Numeric 样式

\journal{Pattern Recognition}

\begin{document}

\begin{frontmatter}

%% --- 标题 ---
\title{\Large CGFformer: Cluster-Guidance Frequency Transformer for Pansharpening}

%% --- 作者信息 (严格遵守单一通讯作者要求) ---

% 作者 1: Zijian Zhou (共同一作)
\author[aff1]{\footnotesize Zijian Zhou\fnref{fn1}}
\ead{zzj210119@stu.xjtu.edu.cn}

% 作者 2: Jianing Zhang (共同一作)
\author[aff1]{\footnotesize Jianing Zhang\fnref{fn1}}
\ead{2518720025@stu.xjtu.edu.cn}

% 作者 3: Kai Sun (唯一通讯作者)
\author[aff2]{\footnotesize Kai Sun\corref{cor1}}
\ead{kaisun@mail.xjtu.edu.cn}

% 作者 4: Xiangyu Zhao
\author[aff2]{\footnotesize Xiangyu Zhao}
\ead{2196410184@stu.xjtu.edu.cn}

\author[aff2]{\footnotesize Chunxia Zhang}
\ead{cxzhang@mail.xjtu.edu.cn}

\author[aff3]{\footnotesize Xiangyong Cao}
\ead{caoxiangyong@mail.xjtu.edu.cn}

%% --- 地址/单位信息 ---
\affiliation[aff1]{organization={College of Artificial Intelligence, Xi’an Jiaotong University},
                city={Xi’an},
                postcode={710049},
                country={China}}

\affiliation[aff2]{organization={School of Mathematics and Statistics, Xi’an Jiaotong University},
                city={Xi’an},
                postcode={710049},
                country={China}}

\affiliation[aff3]{organization={School of Computer Science and Technology, Xi’an Jiaotong University},
                city={Xi’an},
                postcode={710049},
                country={China}}

%% --- 脚注内容 ---
\cortext[cor1]{Corresponding author.}
\fntext[fn1]{These authors contributed equally to this work.}

%% --- 摘要 ---
\begin{abstract}
Pansharpening aims to generate high-resolution multispectral (HRMS) images by fusing low-resolution multispectral (LRMS) images with high-resolution panchromatic (PAN) images. However, the current mainstream frequency-based pansharpening methods employ fixed frequency filters, which cannot precisely adapt to complex and spatially diversified frequency distributions in PAN and MS images. Furthermore, existing denoising strategies insufficiently exploit frequency components for denoising and struggle to suppress various noise types accurately.
To address these challenges, we propose CGFformer, a cluster-guidance frequency Transformer that focuses on varying frequency distribution and interactions between frequency and spatial components.
Specifically, we design an adaptive separation module that integrates local features and non-local information through K-means clustering, enabling more precise separation of high- and low-frequency components.
Subsequently, we introduce a dual-stream refinement module combined with Transformer-based cross-attention to remove various noise, allowing the network to jointly suppress frequency-relevant and irrelevant disturbances.
In addition, we develop a frequency-spatial fusion module designed to enhance detail and facilitate spatial-frequency interaction, ensuring more effective reconstruction of spatial structures in the fused results.
Extensive experiments on multiple benchmark datasets demonstrate that the proposed CGFformer achieves notable improvements over existing pansharpening approaches. 
\end{abstract}

%% --- 关键词 ---

\end{frontmatter}

%\pagenumbering{arabic}

\section{Introduction}

Clear remote sensing images and advanced remote sensing technology play a vital role in fields such as precise nature monitoring \cite{ref2}, urban planning \cite{ref4} and ecological protection \cite{ref5}. However, the existing technologies can only obtain low-resolution multispectral (LRMS) images and high-resolution panchromatic (PAN) images. Among them, LRMS images contain four, eight or more channels in different bands. The spatial resolution of PAN images is usually four times higher than that of LRMS images, but PAN images lack spectral information because they are single-band grayscale images. To solve this problem, pansharpening, which fuses LRMS images with PAN images, is introduced.

Existing pansharpening methods can be divided into traditional methods and deep learning-based methods \cite{ref6}. Specifically, traditional pansharpening methods can be divided into three categories \cite{ref7}: component substitution (CS) methods \cite{ref8}, multiresolution analysis (MRA) methods \cite{ref11}, and variational optimization (VO) methods \cite{ref14}. The core of CS methods is separation and substitution to achieve the generation of HRMS images; however, this type of method is also prone to spectral distortion due to differences in spectral features \cite{ref17}. MRA methods achieve pansharpening through extraction and injection, but they are easy to cause spatial detail distortion in the process \cite{ref18}. VO methods are realized by minimizing the energy function, but when the actual data deviate from the preset model, the problem of spectral shift will occur \cite{ref19}.

Deep learning has achieved remarkable progress in the field of image processing, and a variety of deep learning methods based on convolutional neural networks (CNNs) have been widely applied in pansharpening tasks \cite{ref20,ref21,ref22}. Early deep learning methods concatenate PAN images and upsampled MS images along the channel dimension, then input the concatenated feature into the neural network to generate HRMS images \cite{ref20}. Subsequently, a series of improved methods based on the above ideas were proposed, which significantly enhanced the fusion quality \cite{ref23,ref25}. Among pansharpening methods based on neural networks, frequency-based image fusion techniques have attracted particular attention. In the network architecture represented by PanNet \cite{ref26}, frequency details of images are separated and extracted using a Gaussian filter to generate HRMS images. However, the frequency details obtained merely through a simple filter are insufficient to support the generation of high-quality HRMS images \cite{ref27}. Therefore, Zhou et al. \cite{ref28,ref29} employed Fourier transform instead of spatial filtering for frequency separation, and pointed out that spatial details in remote sensing images are usually associated with high-frequency components, whereas global structures and spectral information are contained in low-frequency components \cite{ref30}. Based on this finding, Diao et al. \cite{ref31} proposed that pansharpening techniques based on frequency separation should effectively address the distribution differences between high- and low-frequency components and constructed HLF-Net to handle high- and low-frequency features by distinct functional modules.

\begin{figure}[t]
  \centering
  \includegraphics[width=1.0\textwidth]{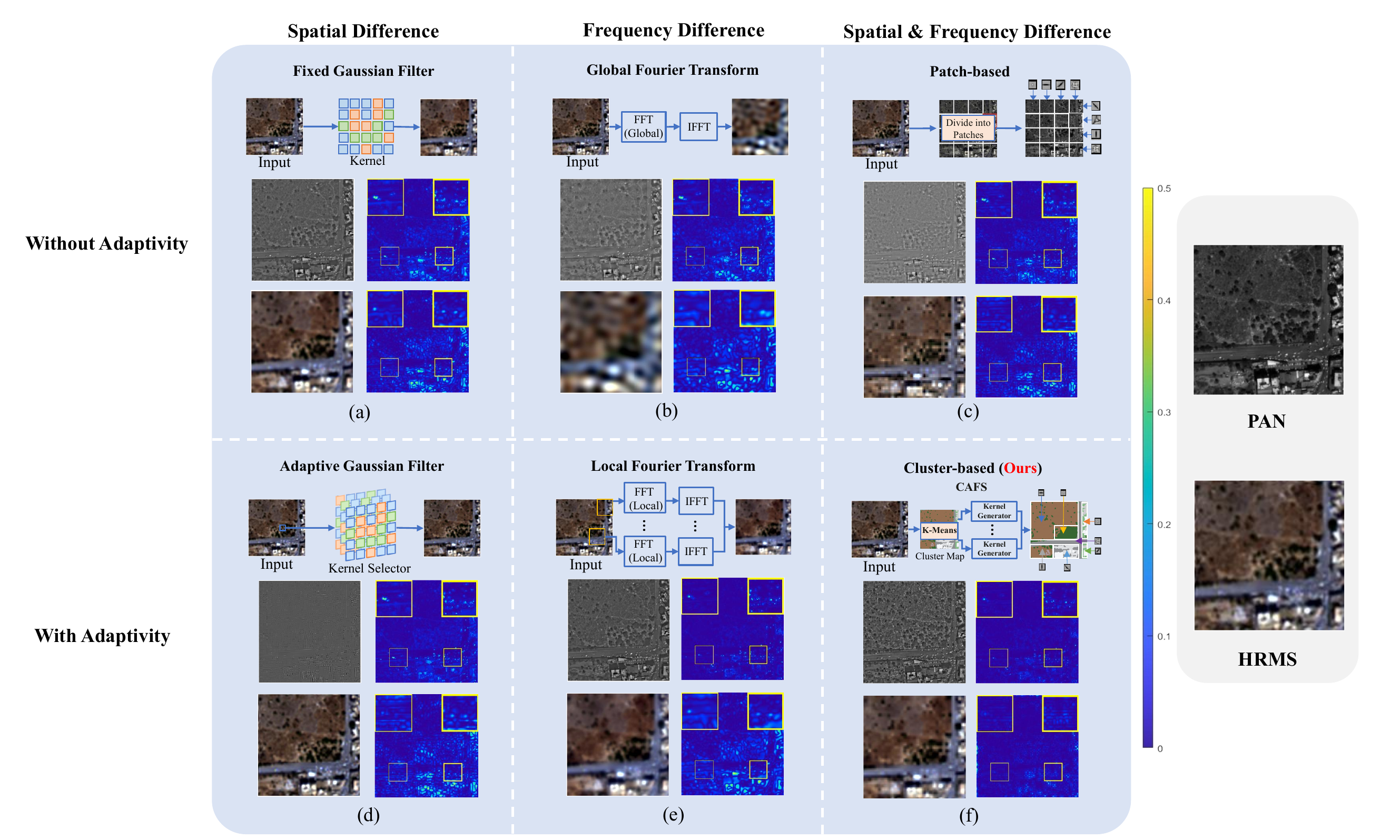}
  \caption{Processes and effects of different methods for frequency separation. The separation mechanisms of different methods are categorized by difference domain (Spatial, Frequency, Spatial \& Frequency) and adaptivity (Without Adaptivity, With Adaptivity). The rightmost panel shows the inputs for separation: the upper is the original PAN image, and the lower is the upsampled MS (HRMS) image. The remaining sections display the corresponding mechanisms and visual results, where each result group contains 4 sub-images (top-left: PAN high-frequency; top-right: PAN high-frequency MAE; bottom-left: HRMS low-frequency; bottom-right: HRMS low-frequency MAE): (a) Fixed Gaussian Filter; (b) Global Fourier Transform; (c) Patch-based; (d) Adaptive Gaussian Filter; (e) Local Fourier Transform; (f) Cluster-based (Ours): Considers non-local similar regions for adaptivity. These results demonstrate the limitations of previous methods and verify the effectiveness of our proposed framework in capturing spatial information in a non-local similar pattern.}
  \label{fig:fig1}
\end{figure}

Although frequency-fusion architectures have achieved notable progress, they still encounter inherent limitations when applied to remote sensing images. Due to the strong coupling between frequency and pixel-level spatial variations, it remains difficult for existing approaches to adapt the diverse frequency distributions present in MS and PAN images \cite{ref32}. Early studies attempted frequency separation through handcrafted filtering. For instance, HLF-Net \cite{ref31} employs Gaussian filters to extract frequency details by modeling spatial differences between image structures, as shown in Fig. \ref{fig:fig1}(a), its fixed and non-adaptive filtering scheme inherently limits the separation capability. In response to this limitation, subsequent works employed various transformation strategies: Zhou et al. \cite{ref33} replaced conventional filters with Fourier transform and performed frequency separation by exploiting frequency discrepancies; Xing et al. \cite{ref34} designed a chunking architecture that integrates spatial differences across hierarchical stages with frequency differences captured by wavelet transform, enabling multi-level frequency decomposition of remote sensing images; and He et al. \cite{ref35} further introduced a DCT-guided kernel selection strategy, which leverages pixel-level discrepancies to achieve frequency separation with pixel-wise adaptivity, as shown in Fig. \ref{fig:fig1}(b)-(d). However, these single-transform operations are fundamentally limited because they cannot sufficiently decouple frequency information from spatially correlated structures, making them inadequate for complex frequency patterns in remote sensing images. To address this problem, Zou et al. \cite{ref32} emphasized that pixel frequency characteristics vary dynamically with spatial locations and thus effective frequency separation should exhibit spatial adaptivity. They proposed a local-neighborhood adaptive strategy, which further inspired the locally adaptive low-pass filter in Liu et al. \cite{ref36} that exploits frequency transformations across spatial regions, as shown in Fig. \ref{fig:fig1}(e). However, Duan et al. \cite{ref37} pointed out that such approaches are inherently restricted to local adaptivity and fail to incorporate non-local similar regions. In other words, an ideal frequency separation mechanism should jointly model both spatial and frequency discrepancies, not only adapt to local regions but also utilize information from non-local similar regions to complete the separation and leverage local and non-local adaptivity simultaneously. Based on this comprehension, as illustrated in Fig. \ref{fig:fig1}(f), our method targeted filter generation with adaptive clustering, enabling abundant local features and non-local similar information and ultimately achieving complementary enhancement in frequency separation.

\begin{figure}[t]
  \centering
  \includegraphics[width=1.0\textwidth]{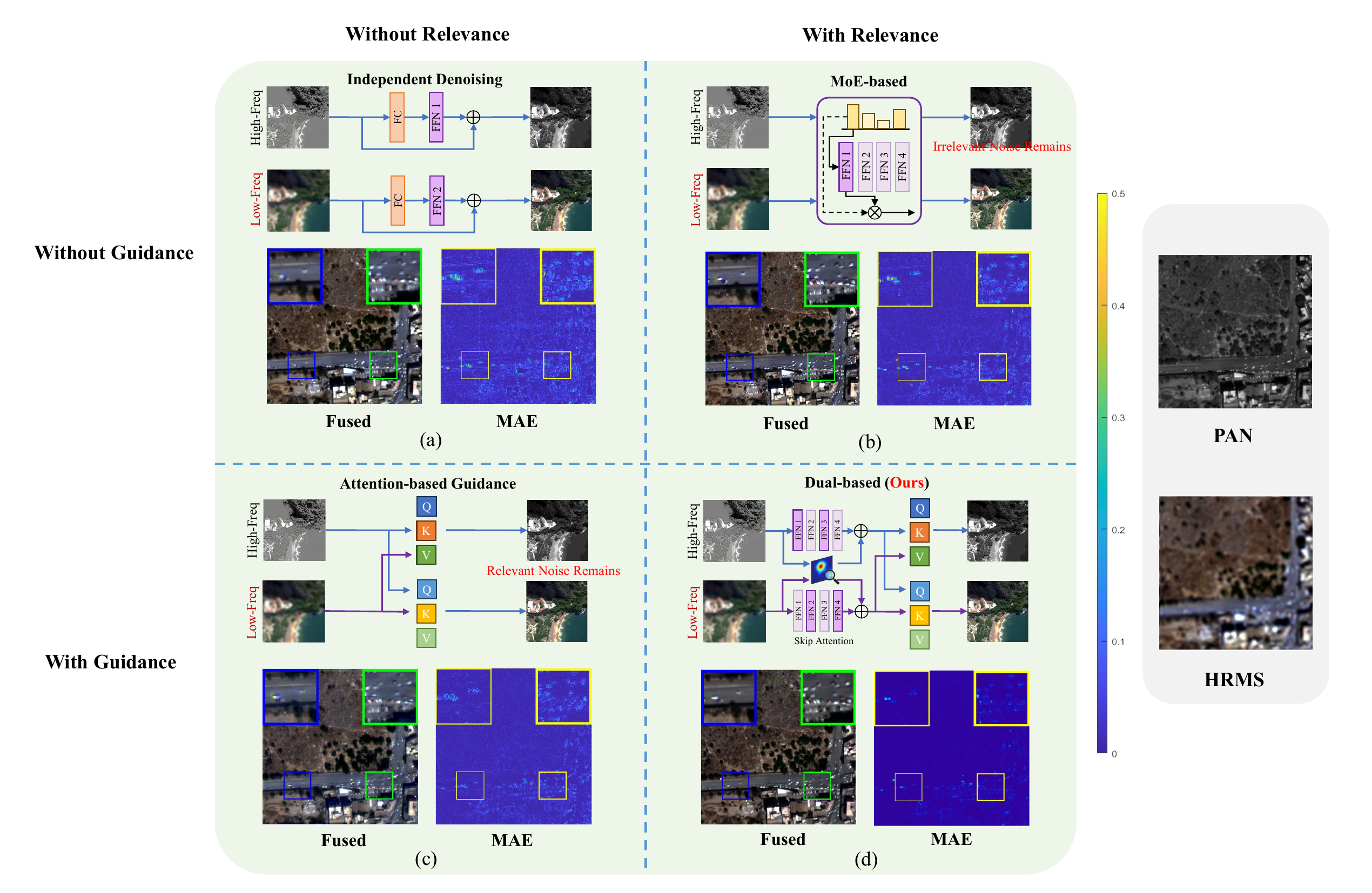}
  \caption{Processes and performance of different frequency-component denoising methods. The denoising mechanisms are categorized by the consideration of frequency relevance (Without Relevance, With Relevance) and cross-frequency guidance (Without Guidance, With Guidance). The rightmost panel shows the inputs for denoising: the upper is the original PAN image, and the lower is the upsampled MS (HRMS) image. The remaining sections illustrate the denoising results of the four methods corresponding to the aforementioned categories, where each result group includes the fused image and its corresponding mean absolute error (MAE) map: (a) Independent Denoising; (b) MoE-based; (c) Attention-based Guidance; (d) Dual-based (Ours): Achieves comprehensive suppression of both frequency-relevant and frequency-irrelevant noise through the proposed framework. Compared to previous methods, our method effectively preserves frequency details while eliminating nonrelevant and relevant noise, demonstrating superior denoising performance.}
  \label{fig:fig2}
\end{figure}

Besides the challenge of frequency separation, the high- and low-frequency components may suffer from noise amplification and artifact generation during decomposition \cite{ref38}, which subsequently hampers frequency-based feature extraction and fusion. Existing studies have explored various strategies for suppressing such degradation. HLF-Net \cite{ref31} performs denoising through single structure filters; however, as shown in Fig. \ref{fig:fig2}(a), the absence of explicit frequency guidance prevents these filters from exploiting the intrinsic characteristics of different frequency components, both frequency-relevant noise embedded within the components and frequency-irrelevant noise introduced during processing can only be weakly suppressed. To enhance flexibility, He et al. \cite{ref35} introduced an MoE-based multi-CNN scheme as shown in Fig. \ref{fig:fig2}(b). Nevertheless, without guidance from frequency components, CNNs with fixed parameters lack the adaptivity required to precisely distinguish and suppress different categories of noise, leading to only marginal attenuation of both frequency-relevant and frequency-irrelevant noise. More recently, Liu et al. \cite{ref36} introduced a Transformer-based guidance mechanism that exploits global low-frequency structures to suppress artifacts in high-frequency components, while preserving feature integrity through high-frequency feedback. Although this strategy improves structural consistency, as illustrated in Fig. \ref{fig:fig2}(c), it mainly targets frequency-irrelevant noise and overlooks frequency-relevant noise that is inherently coupled with the frequency components themselves \cite{ref100}. These observations expose a fundamental limitation of existing approaches: an effective frequency-based denoising strategy should not only exploit the intrinsic characteristics of each frequency component, but also leverage component interactions to guide the targeted suppression of different noise types. Building upon this insight, as shown in Fig. \ref{fig:fig2}(d), our method integrates targeted denoising with frequency guidance, enabling accurate removal of both frequency-relevant and frequency-irrelevant noise and ultimately achieving complementary enhancement across frequency components.

In addition, in the field of frequency fusion, Bandara et al. \cite{ref39,ref40} employ the overcomplete theory to control the receptive field, yet they over-rely on convolutional blocks and repeated upsampling mechanism to enhance the detail preservation effect, which ultimately restricts the model’s frequency fusion capability. Moreover, due to the particularity of remote sensing images, relying solely on frequency features tends to cause information blurring and spatial information loss \cite{ref41}. Thus neural networks that do not utilize spatial features exhibit limited capability to capture finer-grained semantic information and more precise spatial information in remote sensing images.

To address the challenges in adaptive frequency separation, noise suppression, and spatial–frequency fusion in pansharpening, we propose a new Transformer-based framework named CGFformer. The method is built upon three core components, each tailored to the characteristics of remote sensing images. First, we introduce a cluster adaptive frequency separation (CAFS) module, which integrates K-Means clustering with CNN-based filter generation. By partitioning feature representations into multiple non-local similarity groups and dynamically producing filters for each group, the module adaptively captures both local neighborhood relations and long-range similarity cues. This design enables more accurate decomposition of high- and low-frequency components while preserving their spatial relevance. Meanwhile, to alleviate the noise and artifacts introduced during frequency decomposition, we develop a dual-stream refinement (DSR) module. It employs a noise estimation network to identify channel-specific noise patterns, suppresses frequency-relevant degradation by a denoising network, and further refines the separated components through a Transformer-based bidirectional guidance mechanism. This ensures effective feature preservation while reducing frequency-irrelevant interference. Moreover, to enhance representation quality and promote more effective interaction, we refine the fusion architecture with two complementary functions: a spatial enhancement (SFA-S) branch designed with lightweight linear layers to strengthen structural detail, and a spatial-frequency fusion (SFA-F) branch in the spatial-frequency attention (SFA) module to integrate frequency cues and spatial semantics in a coherent manner. Together, these components promote robust collaboration across spatial and frequency domains and ensure high quality reconstruction of HRMS images.

Overall, our contributions are summarized as follows:
\begin{enumerate}
    \item We propose CGFformer, a new pansharpening framework that integrates spatially adaptive frequency separation, frequency-guidance denoising, and spatial-frequency fusion. The model addresses key limitations of existing frequency-based methods by introducing an adaptive separation strategy and a more reliable mechanism for suppressing noise in both high- and low-frequency components. Extensive experiments show that CGFformer consistently surpasses state-of-the-art approaches, providing a unified and effective solution for frequency pansharpening and denoising.
    \item We design a content-adaptive non-local (CAN) block in cluster adaptive frequency separation (CAFS) module that jointly exploits local neighborhoods and non-local similarity while leveraging CNN-based filter generation. This enables accurate and robust discrimination between high- and low-frequency components. In addition, the DSR module utilizes frequency-guidance attention to selectively suppress noise and artifacts while retaining essential structures. Experiments demonstrate the strong generality of these modules across frequency-oriented pansharpening scenarios.
    \item We propose a spatial-frequency fusion mechanism composed of a spatial enhancement (SFA-S) branch and a spatial-frequency fusion (SFA-F) branch. The first branch strengthens spatial detail representation, while the other branch integrates enhanced spatial cues with frequency information in a coherent and efficient manner. Together, they ensure accurate propagation and effective learning of spatial-frequency representations within the network.
\end{enumerate}

\section{Related Work}
\subsection{CNN-Based Pansharpening}
Deep learning has become a dominant paradigm in pansharpening and has demonstrated substantial advantages over traditional model-based approaches. Following the seminal success of the super-resolution network SRCNN \cite{ref42}, Masi et al. \cite{ref20} introduced PNN, which applied convolutional neural networks to learn the nonlinear relationship between PAN and MS images. Although PNN marked an early improvement, its representational capacity remains limited \cite{ref26}. To enhance spatial detail transfer, He et al. \cite{ref23} proposed directly injecting PAN images derived spatial features into MS images, a strategy that has since evolved into a widely adopted framework in pansharpening pipelines.

Despite these advancements, the intrinsic complexity of remote sensing scenes such as cluttered backgrounds, dense small objects, and occlusions, still constrains the effectiveness of purely CNN-based detail injection. To mitigate these challenges, dual-scale learning has emerged as another research direction. P2Sharpen \cite{ref43} leverages a scale-adapted loss function to improve deep pansharpening performance, while ZS-Pan \cite{ref44} further extends this idea by employing the training strategy that simplifies both optimization and inference while improving generalization. However, existing dual-scale frameworks make them highly sensitive to the design of loss functions and associated hyperparameters, which not only complicates tuning but also limits the robustness of these methods in diverse operating scenarios.

\subsection{Transformer-Based Pansharpening}
To address the intrinsic locality limitation of convolutional neural networks, attention\-/based models have been increasingly explored in pansharpening due to their ability to capture long-distance feature dependencies. Meng et al. \cite{ref45} introduced a Transformer backbone into the fusion pipeline, demonstrating the feasibility of global context modeling for detail enhancement. Zhou et al. \cite{ref46} further adopted the Swin-Transformer architecture \cite{ref47} to reduce parameter complexity while maintaining competitive reconstruction accuracy. Although these Transformer-based approaches achieve notable improvements, they generally struggle to effectively preserve and reconstruct high-frequency information, which remains critical for spatial detail fidelity in pansharpening.

Bandara et al. \cite{ref48} proposed HyperTransformer, which employs independent feature extraction branches and multi-scale Transformer modules tailored to inputs of different resolutions. While the model delivers high-quality fusion results, its architecture depends heavily on convolution blocks and repeated upsampling operations, ultimately limiting its representational capacity and hindering its ability to capture spatial structures.

Meanwhile, Liu et al. \cite{ref36} introduced FSGformer, which incorporates a multi-window mechanism and a spectral-spatial guidance strategy to better adapt Transformer modeling to multispectral characteristics \cite{ref49}. Despite its advantages, the guidance mechanism may reintroduce coupling effects between branches, potentially degrading overall fusion performance under complex image conditions.

\subsection{Pansharpening Based On Frequency Fusion}
Frequency-based fusion has become an important research direction in pansharpening. Early works by Yang et al. \cite{ref26} and Diao et al. \cite{ref31} employed high-frequency filtering to extract detail information for fusion. With further advancements, Zhou et al. \cite{ref28}, \cite{ref29} examined image fusion from both spatial and frequency perspectives and incorporated invertible neural networks to build a unified fusion framework. However, frequency transforms with a single formulation cannot adequately model the diverse and spatially varying frequency distributions present in remote sensing images, which limits their adaptability. In recent research, wavelet-based methods have attracted extensive attention. Xing et al. \cite{ref34} applied wavelet transform with MRA to optimize frequency-domain fusion. Huang et al. \cite{ref53} combined wavelet transform with multi-scale spatial enhancement to jointly handle spatial and frequency features, leading to improved fusion accuracy. Nevertheless, wavelet decomposition inherently compromises the scale integrity of frequency components and lose spatial precision, making it difficult to retain structural details in the reconstruction.

Other studies target the intrinsic frequency differences between PAN and MS images and adopt frequency separation strategies prior to fusion. Diao et al. \cite{ref31} used handcrafted filters to isolate high- and low-frequency components and trained them independently. Liu et al. \cite{ref36} introduced a locally adaptive filter and performed frequency-specific learning with attention mechanisms to enhance band-wise feature extraction. Frequency separation methods grounded solely in local spatial neighborhoods tend to overemphasize local variations while neglecting non-local structural similarity, resulting in incomplete or suboptimal frequency decomposition. To address this problem, frequency separation is formulated by integrating local neighborhood features with non-local similar region information in our work to enhance the accuracy of frequency component.

\section{Proposed Method}
\begin{figure}[t]
  \centering
  \includegraphics[width=1.0\textwidth]{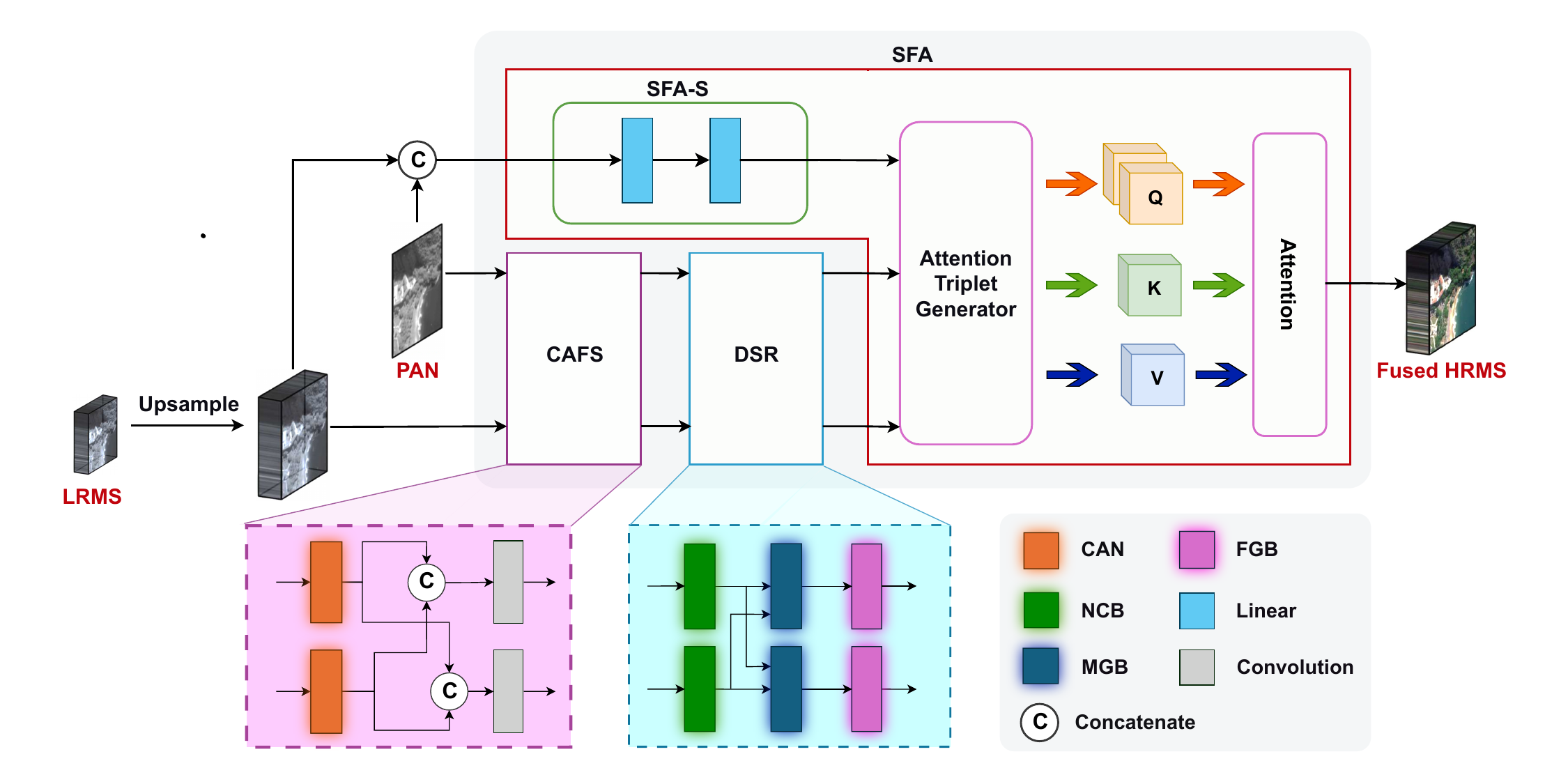}
  \caption{CGFformer network architecture diagram. Our CGFformer network is mainly composed of three parts: the CAFS module for separating high- and low-frequency components, as shown in Fig. \protect\ref{fig:fig4}; the DSR module for suppressing dual-type noise, and the SFA module for fusing frequency components while injecting spatial information, as shown in Fig. \protect\ref{fig:fig5}. CAN is the content-adaptive non-local block. NCB is the noise calibration block, MGB is the mutual guidance block and FGB is the feature gating block.}
  \label{fig:fig3}
\end{figure}

This section elaborates on the overall architectural design of the proposed CGFformer model. This model is mainly composed of three core modules: the cluster adaptive frequency separation (CAFS) module, the dual-stream refinement (DSR) module, and the spatial-frequency attention (SFA) module. The overall workflow of CGFformer is shown in Fig. \ref{fig:fig3}.

In this study, the input and output data of the model are defined as follows: let the input PAN images be $\bm{I}_P\in \mathbb{R}^{H\times W\times1}$, the MS images be $\bm{I}_M\in \mathbb{R}^{\left(H/r\right)\times\left(W/r\right)\times C}$, and the GT images be $\bm{I}_H\in \mathbb{R}^{H\times W\times C}$, Where $H\times W$ represents the spatial resolution of the PAN images, $r$ is the reduction rate, and $C$ is the number of spectral channels. Based on the above definition, the core optimization objective of the CGFformer network can be expressed as follows:

\begin{equation}\label{eq:loss_function}
\begin{split}
\min_{\theta} \mathcal{L}( &SFA\left(DSR\left(CAFS\!\left(\bm{I}_P, \bm{I}_M\uparrow\right)\right),\right.\bm{I}_H; \theta)
\end{split}
\end{equation}

\noindent \lowercase{w}here $\mathcal{L}$ represents the total loss function of the network, and $\uparrow$ represents bilinear interpolation upsampling. CAFS, DSR and SFA correspond to the three core modules proposed in this study.

\subsection{CAFS: Cluster Adaptive Frequency Separation Module}
\begin{figure}[t]
  \centering
  \includegraphics[width=1.0\textwidth]{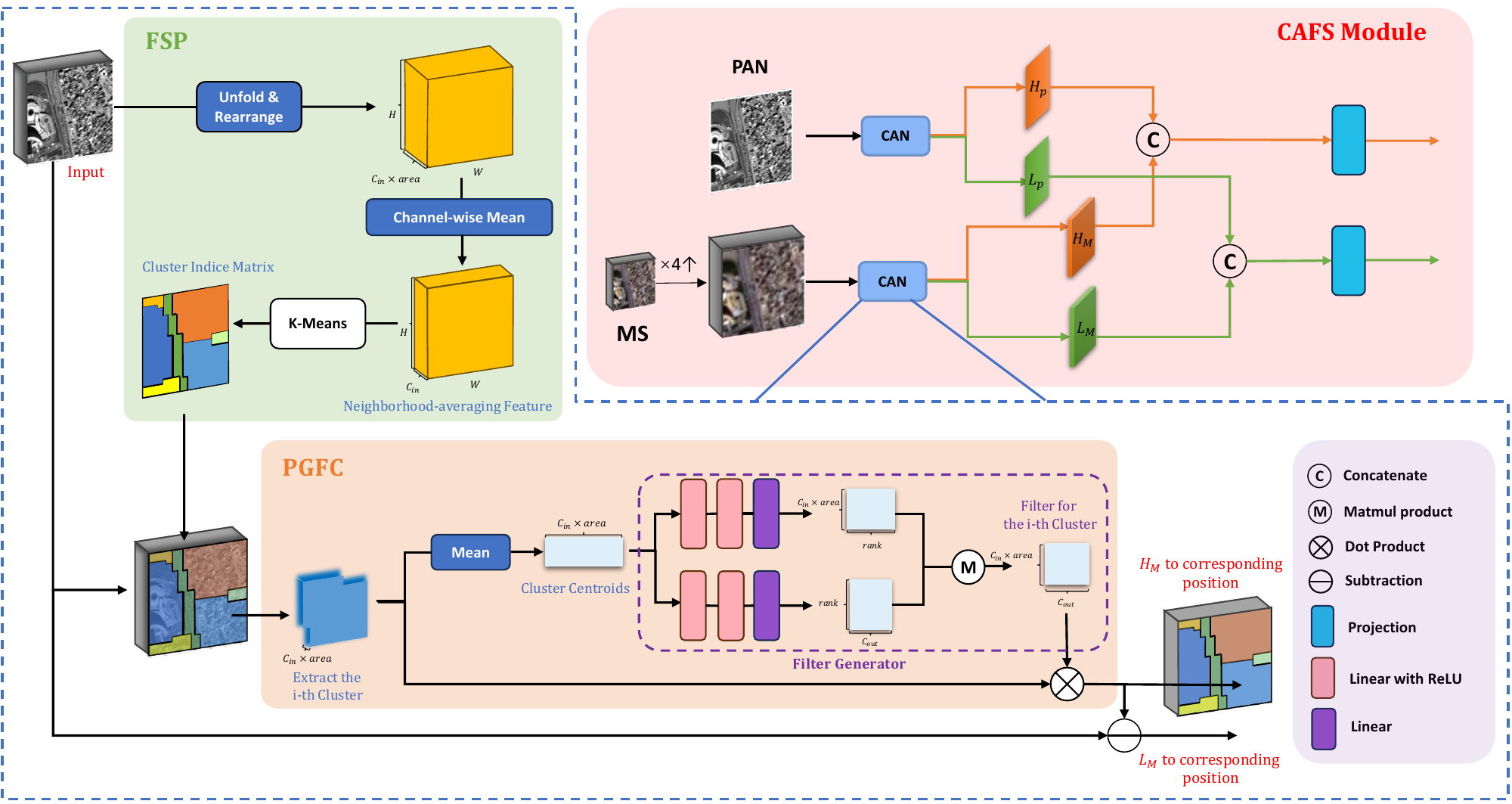}
  \caption{Details of the CAFS module. The upper right part shows the overall structure diagram of the CAFS module, where CAN stands for the content-adaptive non-local block. The left and lower parts present the details of the CAN block.}
  \label{fig:fig4}
\end{figure}

In order to achieve the accurate separation of high- and low-frequency components of remote sensing images, this paper first designs the CAFS module as the fundamental support. Zou et al. \cite{ref32} pointed out that the frequency characteristics of pixels in remote sensing images are location-dependent, that is, the frequency information will be dynamically adjusted as the spatial position of pixels changes. Furthermore, inspired by the concept of non-local similarity modeling proposed by Duan et al. \cite{ref37}, we propose a cluster adaptive frequency separation (CAFS) module, which can adaptively and accurately separate the frequency components of images by leveraging non-local similarity clustering and location-sensitive filtering mechanism. The detailed workflow of the CAFS module is shown in Fig. \ref{fig:fig4}.

\textit{1) CAN: Content-Adaptive Non-Local Block. }Firstly, the upsampled MS image $\bm{I}_M\uparrow$ and the PAN image $\bm{I}_P$ are input into the CAN block, respectively, which separates the high- and low-frequency components. The CAN block mainly consists of two parts: a feature similarity-based partition (FSP) sub-module and a partition-guided filter calculation (PGFC) sub-module. Unlike traditional fixed filters that apply a uniform kernel across the entire image, this partition-guided design allows the network to perceive spatially diversified frequency distributions. These two parts work together to implement the frequency separation function.

\textbf{The feature similarity-based partition (FSP) sub-module} is implemented as follows: For any pixel position $\left(x,y\right)$ in the image, a $k\times k$ sliding window is used to extract the neighborhood features $f_{x,y}$ of the pixel. In Fig. \ref{fig:fig4}, the term $area$ denotes the spatial expanse of this window, defined as $area=k \times k$. The calculation process is given in the following formula:

\begin{equation}
f_{x,y} = \frac{1}{k^2} \sum_{x'=-\lfloor k/2 \rfloor}^{\lfloor k/2 \rfloor} \sum_{y'=-\lfloor k/2 \rfloor}^{\lfloor k/2 \rfloor} \bm{I}_{x+x', y+y'}
\end{equation}

\noindent \lowercase{w}here $f_{x,y}\in \mathbb{R}^C$, the variable $\bm{I}$ in the equation can refer to $\bm{I}_M\uparrow$ or $\bm{I}_P$. Based on the above-extracted set of neighborhood feature vectors $\{f_{x,y}\}_{x=1,y=1}^{H,W}$ of the entire image, the K-Means clustering algorithm \cite{ref54} is employed to partition all pixels of the image. Unlike convolution operations confined to local windows, this clustering strategy aggregates widely separated but semantically similar regions, thereby effectively capturing the non-local similarity cues required to differentiate complex frequency patterns. Finally, a clustering index matrix $\bm{I}_{cls}\in \mathbb{N}^{H\times W}$ is obtained, where $\bm{I}_{cls}\left(x,y\right)$ represents the cluster number to which the pixel at position $\left(x,y\right)$ belongs, and satisfies $1\le \bm{I}_{cls}\left(x,y\right)\le K$ (where $K$ is the preset number of clusters).

\textbf{The partition-guided filter calculation (PGFC) sub-module} is as follows. First, the pixel set corresponding to the $i$-th cluster is defined.

\begin{equation}
\bm{S}_i=\left\{\left(x,y\right)\mid \bm{I}_{cls}\left(x,y\right)=i\right\}
\end{equation}

To characterize the global frequency properties of each cluster, the centroid vector of the neighborhood features of all pixels within cluster $\bm{S}_i$ is computed, the calculation of the centroid vector $c_i$ is given in the following formula:

\begin{equation}
c_i = \frac{1}{|\bm{S}_i|} \sum_{(x,y) \in S_i} \bm{f}_{x,y}
\end{equation}

\noindent \lowercase{w}here $\left|\bm{S}_i\right|$ denotes the total number of pixels in cluster $\bm{S}_i$, and $c_i\in \mathbb{R}^{k^2\times C}$.

Then, a position-adaptive low-pass filter $\bm{W}_i$ is designed based on the cluster centroid vector $c_i$. To reduce the filter’s parameter redundancy, a low-rank decomposition strategy is employed: two independent yet structurally identical MLP blocks are designed, denoted as $\mathrm{mlp}_A$ and $\mathrm{mlp}_B$ respectively. $\mathrm{mlp}_A$ maps the centroid vector $c_{i}$ to a matrix $\bm{A}$ with dimensions $(C_{in}\times k^{2})\times rank$, while $\mathrm{mlp}_B$ maps the centroid vector $c_{i}$ to a matrix $\bm{B}$ with dimensions $rank\times C_{out}$. Here, $rank$ represents the hyperparameter controlling the matrix decomposition rank, while $C_{in}$ and $C_{out}$ denote the number of input and output feature channels respectively.

Using the generated cluster-specific filter $\bm{W}_i$ through matrix multiplication between $\bm{A}$ and $\bm{B}$, targeted frequency separation for $\bm{I}_M\uparrow$ and $\bm{I}_P$ is realized. The mathematical expressions are given in the following formula:

\begin{subequations}
\begin{align}\label{eq:mlp_ab}
& \bm{L}_M(x,y) = \bm{W}_{I_{cls}(x,y)} \otimes \bm{f}_{x,y} \\
& \bm{H}_M = \bm{I}_M\uparrow - \bm{L}_M 
\end{align}
\end{subequations}

Similarly, for PAN images $\bm{I}_P$, the calculation logic is consistent with that of $\bm{I}_M\uparrow$.

\begin{subequations}\label{eq:mlp_a}
\begin{align}
& \bm{L}_P(x,y) = \bm{W}_{I_{cls}(x,y)} \otimes \bm{f}_{x,y} \\
& \bm{H}_P = \bm{I}_P - \bm{L}_P
\end{align}
\end{subequations}

In the above formulas, $\otimes$ denotes the matrix multiplication operation, $\bm{L}_M$ and $\bm{L}_P$ correspond to the low-frequency components of $\bm{I}_M\uparrow$ and $\bm{I}_P$, $\bm{H}_M$ and $\bm{H}_P$ correspond to their high-frequency components.

\textit{2) Projection Block. }To unify feature dimensions and adapt to subsequent module input requirements, these features are concatenated along the channel dimension, and then mapped to the same feature space via a $3\times 3$ convolutional layer with shared parameters.

\begin{subequations}\label{eq}
\begin{align}
\bm{H}_E &= \mathrm{Proj}\!\left(\mathrm{Concat}\!\left(\bm{H}_P, \bm{H}_M\right)\right) \\
\bm{L}_E &= \mathrm{Proj}\!\left(\mathrm{Concat}\!\left(\bm{L}_P, \bm{L}_M\right)\right)
\end{align}
\end{subequations}

\noindent \lowercase{w}here $\bm{H}_E$ and $\bm{L}_E$ are the high- and low-frequency features to be input into the subsequent module.

\subsection{DSR: Dual-Stream Refinement Module}

\begin{figure}[t]
  \centering
  \includegraphics[width=1.0\textwidth]{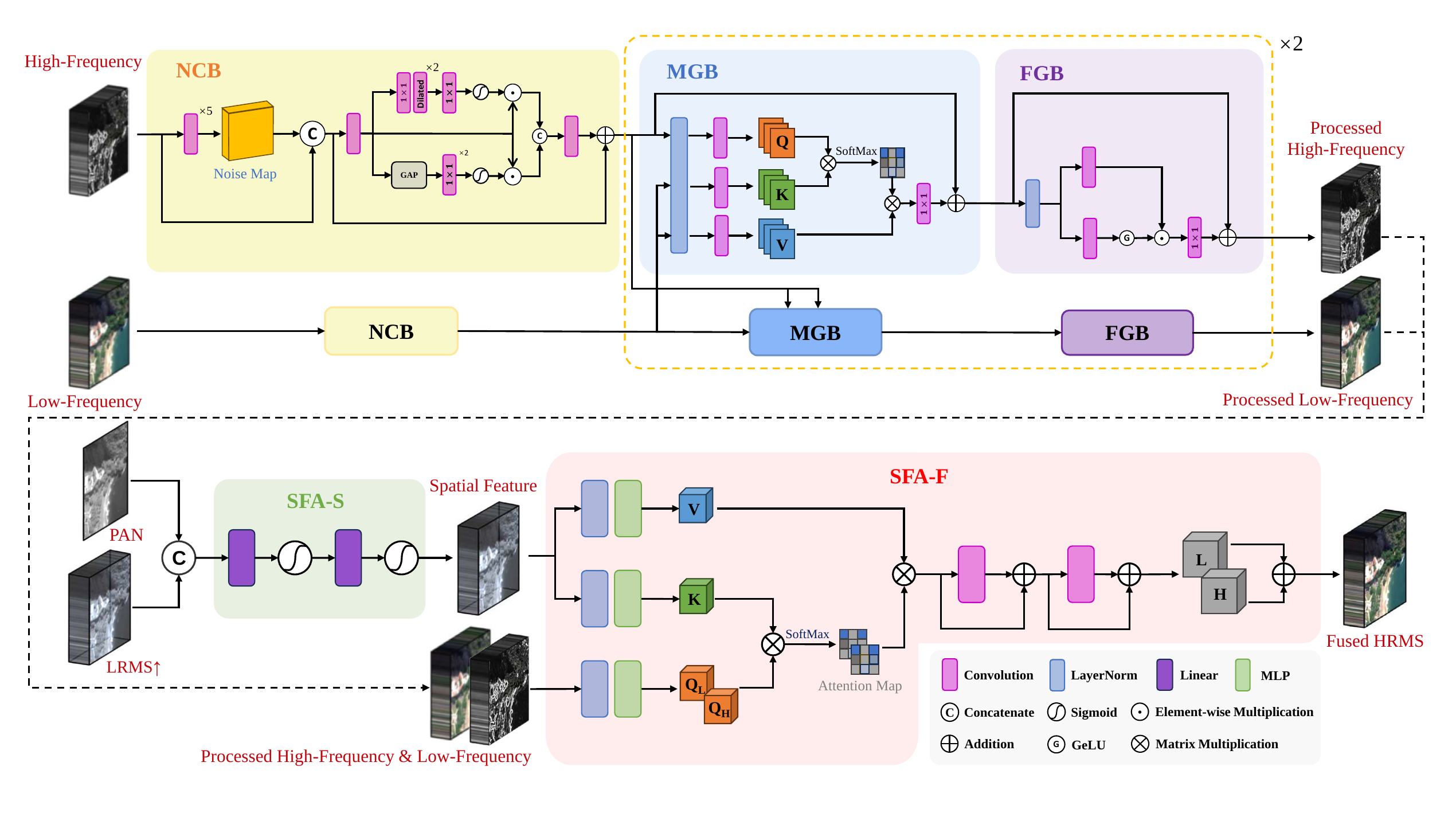}
  \caption{Details of the DSR and SFA module. The upper part shows the overall structure diagram of the DSR module and the lower part shows the SFA module. The whole part shows details of each block in the modules: NCB (Noise Calibration Block), MGB (Mutual Guidance Block), FGB (Feature Gating Block), SFA-S (Space Branch in SFA) and SFA-F (Fusion Branch in SFA).}
  \label{fig:fig5}
\end{figure}

The high- and low-frequency components separated by the CAFS module inevitably contain two types of noise: random noise independent of image content and frequency-relevant noise that overlaps with the effective image frequency distribution. Inspired by the work of Mo et al. \cite{ref100}, we design the DSR module, which adopts a cascaded denoising and optimization strategy. Specifically, a Noise Calibration Block (NCB) is first employed to suppress frequency-relevant noise, followed by two successive stages of cross-attention guidance and feature-gated optimization to progressively eliminate the remaining content-independent random noise and optimize frequency features. The overall workflow of the DSR module is illustrated in the upper part of Fig. \ref{fig:fig5}.

\textit{1) NCB: Noise Calibration Block. } It consists of two steps: a noise estimation step and a noise removal step. First, taking the high-frequency features $\bm{H}_E$ and low-frequency features $\bm{L}_E$ output by the CAFS module as inputs, the noise estimation step utilizes a network structure composed of multi-layer convolutions. The outputs of this network are two noise maps with the same dimensions as the input features: one is $\bm{\sigma}_H$, representing the noise distribution inherent to the high-frequency features, and the other is $\bm{\sigma}_L$, characterizing the noise distribution of the low-frequency features. Finally, these noise maps and the original noisy features are concatenated along the channel dimension to obtain $\bm{H}_C$ and $\bm{L}_C$, which contain the signal noise estimation maps. The noise removal step employs convolutions to extract features from the outputs $\bm{H}_C$ and $\bm{L}_C$ of the noise estimation step to obtain the intermediate feature $\bm{F}_C$. Subsequently, $\bm{F}_C$ is fed into two branches—spatial attention and channel attention—to generate weight maps, which are then used to complete the denoising process.

The spatial attention branch employs dilated convolution blocks to generate the spatial weight map $\bm{M}_S$. The channel attention branch adopts Global Average Pooling (GAP) and an excitation mechanism to generate the channel weight map $\bm{M}_C$. Afterwards, element-wise multiplication is performed between $\bm{F}_C$ and both $\bm{M}_S$ and $\bm{M}_C$. These two weighted features are then concatenated along the channel dimension. After convolutional processing, the final denoised frequency component $\bm{I}_{\mathrm{denoised}}$ is obtained. The specific calculation formulas are as follows:

\begin{subequations}
\begin{align}
& \bm{M}_S = \sigma\left(\mathrm{Conv}_{\mathrm{dilated}}\left(\bm{F}_C\right)\right) \\
& \bm{M}_C = \sigma\left(\mathrm{Conv}\left(\frac{1}{H\times W}\sum_{x=1}^{H}\sum_{y=1}^{W}\bm{F}_C(x,y,:)\right)\right) \\
&\bm{I}_{\mathrm{denoised}} = \mathrm{Conv}\left( \mathrm{Concat}\left(\bm{F}_C \odot \bm{M}_S, \bm{F}_C \odot \bm{M}_C\right) \right)
\end{align}
\end{subequations}

Here, $\sigma$ represents the sigmoid function, $\odot$ denotes element-wise matrix multiplication, and $\bm{I}_{\mathrm{denoised}}$ represents the frequency component processed by the NCB module to remove signal-related noise. In the low-frequency branch, $\bm{I}_{\mathrm{denoised}}$ is equivalent to $\bm{L}_{\mathrm{denoised}}$, while in the high-frequency branch, $\bm{I}_{\mathrm{denoised}}$ is equivalent to $\bm{H}_{\mathrm{denoised}}$.

\textit{2) MGB: Mutual Guidance Block. }At this stage, the mutual guidance block is introduced to suppress the remaining random noise.

To construct the input for the mutual attention mechanism, the denoised high-frequency and low-frequency features are first transformed into query (Q), key (K), and value (V) via the feature mapping function. The structure of 1×1 point convolution and $3\times3$ depth convolution is employed in the query mapping function $f_q$, key mapping function $f_k$, and value mapping function $f_v$; layer normalization (LN) is applied to the features to stabilize training prior to mapping.

\begin{subequations}\label{eq:Qhl}
\begin{gather}
\bm{Q}_h\!=\!f_q\!\left(\mathrm{LN}\!\left(\bm{H}\right)\right)\!,\!
\bm{K}_h\!=\!f_k\!\left(\mathrm{LN}\!\left(\bm{L}\right)\right)\!,\!
\bm{V}_h\!=\!f_v\!\left(\mathrm{LN}\!\left(\bm{L}\right)\right)\\
\bm{Q}_l\!=\!f_q\!\left(\mathrm{LN}\!\left(\bm{L}\right)\right)\!,
\bm{K}_l\!=\!f_k\!\left(\mathrm{LN}\!\left(\bm{H}\right)\right)\!,
\bm{V}_l\!=\!f_v\!\left(\mathrm{LN}\!\left(\bm{H}\right)\right)
\end{gather}
\end{subequations}

\noindent where $\bm{H}$ represents $\bm{H}_{\mathrm{denoised}}$, $\bm{L}$ represents $\bm{L}_{\mathrm{denoised}}$.

The mutual attention guidance is divided into two parts: "high-frequency guided low-frequency" and "low-frequency guided high-frequency". In the high-frequency guided low-frequency stage, it takes $\bm{H}_{\mathrm{denoised}}$ as the query $\bm{Q}_h$, and $\bm{L}_{\mathrm{denoised}}$ as the key $\bm{K}_h$ and value $\bm{V}_h$. In the low-frequency guided high-frequency stage, it takes $\bm{L}_{\mathrm{denoised}}$ as the query $\bm{Q}_l$, and $\bm{H}_{\mathrm{denoised}}$ as the key $\bm{K}_l$ and value $\bm{V}_l$. The dimension of the guided features is adjusted via residual connections and $1\times1$ convolutions to obtain the preliminarily optimized high-frequency features $\bm{H}_G$ and low-frequency features $\bm{L}_G$.

\begin{subequations}\label{Attn} % 语义化标签，方便引用
\begin{gather}
\mathrm{Attn}\left(\bm{Q},\bm{K},\bm{V}\right)=\mathrm{SoftMax}\left(\frac{\bm{Q}\bm{K}^T}{\sqrt D}\right)\bm{V}\\
\bm{H}_G=\bm{H}_{\mathrm{denoised}}+{\mathrm{Conv}}\left(\mathrm{Attn}\left(\bm{Q}_l,\bm{K}_l,\bm{V}_l\right)\right)\\
\bm{L}_G=\bm{L}_{\mathrm{denoised}}+{\mathrm{Conv}}\left(\mathrm{Attn}\left(\bm{Q}_h,\bm{K}_h,\bm{V}_h\right)\right)
\end{gather}
\end{subequations}

\textit{3) FGB: Feature Gating Block. }Inspired by the research of Zamir et al. \cite{ref100}, a feature gating block is introduced to suppress redundant information. This network implements the gating mechanism through the GELU activation function and element-wise multiplication $\odot$.

\begin{subequations}\label{eq:fgdf} % 语义化标签，方便引用
\begin{gather}
\mathrm{GAT}(\bm{X})=\mathrm{GELU}\bigl(f_1\bigl(\mathrm{LN}\bigl(\bm{X}\bigr)\bigr)\bigr)\odot f_1\bigl(\mathrm{LN}\bigl(\bm{X}\bigr)\bigr)\bigr)\\
\mathrm{FGB}\bigl(\bm{X}\bigr) = \bm{X} + \mathrm{Map}(\mathrm{GAT}(\bm{X}))\\
\bm{H}_{BG}, \bm{L}_{BG}=\mathrm{FGB}(\bm{H}_G, \bm{L}_G)
\end{gather}
\end{subequations}

The preliminarily optimized $\bm{H}_G$ and $\bm{L}_G$ are input into the FGB network, and finally output the high-frequency features $\bm{H}_{BG}$ and low-frequency features $\bm{L}_{BG}$ optimized via bidirectional guidance and gating. This MGB–FGB pair constitutes one refinement stage, and the same structure is applied twice in DSR to achieve progressive bidirectional optimization. Through this cascaded optimization, the DSR module effectively eliminates not only the random noise unrelated to frequency but also the frequency-relevant noise, ensuring clean and robust inputs for the subsequent fusion stage.

\subsection{SFA: Space-Frequency Attention Module}

To mitigate the issues of information blurring and spatial information loss caused by relying solely on frequency features, the proposed SFA module functions as a fusion engine. First, it explicitly incorporates a Spatial Enhancement (SFA-S) branch and a Space-Frequency Fusion (SFA-F) branch to inject rich spatial details into the frequency domain, ensuring the preservation of geometric structures. Then, serving as the final reconstruction stage, it effectively synthesizes the independently optimized high- and low-frequency components to generate the final HRMS output. The specific workflow of this module is illustrated in the lower part of  Fig. \ref{fig:fig5}.

The core objective of the Spatial Enhancement (SFA-S) branch is to enhance the spatial detail representation of the image within the spatial domain, thereby providing high-quality enhanced spatial feature inputs. Inspired by Huang et al. \cite{ref53}, SFA-S employs two MLP blocks to generate $\bm{F}_S$. The overall process is formulated as follows:

\begin{subequations}
\begin{align}
& \mathrm{SE}(\cdot) = \sigma\left(\mathrm{Linear}(\cdot)\right) \\
& \bm{F}_S = \mathrm{SE}\left(\mathrm{SE}\left(\mathrm{Concat}\left(\bm{I}_P, \bm{I}_M\uparrow\right)\right)\right)
\end{align}
\end{subequations}

Inspired by the study of Yang et al. \cite{ref41}, and considering the difficulty for a single network to fully learn the fine-grained structures of high-frequency features, the Space-Frequency Fusion (SFA-F) step within the SFA module utilizes the outputs $\bm{H}_{BG}$ and $\bm{L}_{BG}$ from the DSR module, along with the output $\bm{F}_S$ from the aforementioned step, to construct a space-frequency attention triplet. To prioritize holistic spatial features within the attention mechanism, the spatial features represented by $\bm{F}_S$ are directly assigned as the Key and Value, $\bm{H}_{BG}$ or $\bm{L}_{BG}$ are assigned as the Query in the triplet. Subsequently, these three components are processed via Layer Normalization (LN) and a Multi-Layer Perceptron (MLP) to enhance their representational capability. The specific process is described by the following equations:

\begin{subequations}
\begin{align}
& \bm{Q}_H, \bm{Q}_L = \mathrm{MLP}\left(\mathrm{LN}\left(\bm{H}_{BG}, \bm{L}_{BG}\right)\right) \\
& \bm{K}, \bm{V} = \mathrm{MLP}\left(\mathrm{LN}\left(\bm{F}_S\right)\right)
\end{align}
\end{subequations}

Afterward, the attention mechanism is utilized to reconstruct the fused features across different frequency components. The results are then processed through MLPs and residual connections to generate the final image containing fused spatial information. The detailed process is as follows:

\begin{subequations}
\begin{gather}
\bm{O}_H = f_I\left(\mathrm{Attn}\left(\bm{Q}_H, \bm{K}, \bm{V}\right)\right) \\
\bm{O}_L = f_I\left(\mathrm{Attn}\left(\bm{Q}_L, \bm{K}, \bm{V}\right)\right) \\
\bm{O} = \bm{O}_H + \bm{O}_L
\end{gather}
\end{subequations}

Here, $f_I$ denotes the operation of applying MLPs and residual connections to the outcome of the multiplication operation, and $\bm{O}\in \mathbb{R}^{H\times W\times C}$ represents the HRMS image generated by the network.

\subsection{Loss Function}
To optimize the proposed network, we employ the $\ell_1$ loss function, which has been widely proven to yield robust and consistent reconstruction performance in pansharpening tasks. The loss function is formulated as follows:
\begin{equation}\label{eq:loss}
    \mathcal{L} = \left\| \bm{O} - \bm{I}_{\text{GT}} \right\|_1
\end{equation}

\noindent where $\bm{O} \in \mathbb{R}^{H \times W \times C}$ denotes the predicted image, and $\bm{I}_{\text{GT}} \in \mathbb{R}^{H \times W \times C}$ represents the corresponding ground truth image.

\section{Experiment}
\subsection{Datasets}
In the field of high-resolution Earth observation, the Gaofen-2 (GF-2) and WorldView\-/3 (WV3) remote sensing datasets hold significant application value. Both of the datasets can provide MS and PAN images, offering high-quality data sources for key tasks in urban planning, resource surveys, environmental monitoring, and other fields. Additionally, these two datasets exhibit distinct characteristics in spatial resolution, spectral band configuration, and imaging swath width, which can meet the requirements of fine-grained and thematic information extraction in different scenarios. Specific parameter information during the experimental process is presented in Table \ref{tab:datasets}.

\begin{table}[t]
\caption{Details of the two datasets (GaoFen2 and WorldView-3) used in this study}
\centering
\fontsize{9.2pt}{10pt}\selectfont
\begin{adjustbox}{center, max width=1.0\textwidth}
\begin{tabular}{>{\centering\arraybackslash}cccccc}
\toprule
Datasets & Train & Reduced-resolution Test & Full-resolution Test \\
\midrule
GaoFen-2 & \makecell{Patches: 19809\\MS: $16 \times 16 \times 4$\\PAN: $64 \times 64 \times 1$} & \makecell{Patches: 20\\MS: $64 \times 64 \times 4$\\PAN: $256 \times 256 \times 1$} & \makecell{Patches: 20\\MS: $128 \times 128 \times 4$\\PAN: $512 \times 512 \times 1$} \\
\cmidrule(lr){1-4}
WorldView-3 & \makecell{Patches: 9714\\MS: $16 \times 16 \times 8$\\PAN: $64 \times 64 \times 1$} & \makecell{Patches: 20\\MS: $16 \times 16 \times 8$\\PAN: $64 \times 64 \times 1$} & \makecell{Patches: 20\\MS: $64 \times 64 \times 8$\\PAN: $256 \times 256 \times 1$} \\
\bottomrule
\end{tabular}
\end{adjustbox}
\label{tab:datasets}
\end{table}

\subsection{Implementation Details}
In the proposed model, the number of clusters in the CAN block is set to 32; the size of the convolution kernel is set to $3 \times 3$; the attention head configuration of the SFA module is set to $8$ . The projection dimension is set to 32. During the training process, the AdamW optimizer is adopted, the batch size is set to 128, and the learning rate is set to $0.0006$. 

\begin{table}[t]
\caption{Quantitative results of the fusion model on the WorldView-3 Reduced-resolution dataset. Bold shows the best result, and italicized values indicate the second-best method. }
\label{tab:tab2}
\centering
\fontsize{9.2pt}{12pt}\selectfont
\setlength{\tabcolsep}{1.95pt}

\newcommand{\tpm}{$\mkern 1mu \pm \mkern 1mu$}

\begin{tabular}{@{} cccccc @{}}
\toprule
\textbf{Method} & \textbf{PSNR↑} & \textbf{SSIM↑} & \textbf{SCC↑} & \textbf{SAM↓} & \textbf{ERGAS↓} \\
\midrule
\textbf{GS} & 29.314\tpm1.927 & 0.851\tpm0.033 & 0.924\tpm0.034 & 6.128\tpm1.782 & 5.524\tpm1.396 \\
\textbf{BDSD} & 30.946\tpm1.407 & 0.888\tpm0.030 & 0.935\tpm0.028 & 5.465\tpm1.671 & 4.654\tpm1.429 \\
\textbf{PRACS} & 29.955\tpm1.377 & 0.856\tpm0.034 & 0.920\tpm0.041 & 5.610\tpm1.674 & 5.200\tpm1.473 \\
\textbf{MTF-GLP} & 31.155\tpm1.377 & 0.892\tpm0.026 & 0.938\tpm0.027 & 5.251\tpm1.557 & 4.553\tpm1.391 \\
\textbf{MF} & 30.412\tpm1.655 & 0.875\tpm0.026 & 0.928\tpm0.028 & 5.317\tpm1.472 & 4.901\tpm1.274 \\
\textbf{PanNet} & 35.084\tpm1.727 & 0.958\tpm0.011 & 0.972\tpm0.026 & 3.793\tpm0.715 & 2.836\tpm0.700 \\
\textbf{FusionNet} & 36.015\tpm1.630 & 0.963\tpm0.010 & 0.976\tpm0.021 & 3.502\tpm0.624 & 2.577\tpm0.584 \\
\textbf{GPPNN} & 34.308\tpm1.542 & 0.953\tpm0.012 & 0.967\tpm0.024 & 4.253\tpm0.803 & 3.081\tpm0.701 \\
\textbf{HLF-Net} & 35.823\tpm1.617 & 0.962\tpm0.010 & 0.976\tpm0.017 & 3.544\tpm0.640 & 2.601\tpm0.590 \\
\textbf{MD$^3$Net} & 36.775\tpm1.641 & 0.969\tpm0.009 & 0.980\tpm0.015 & 3.150\tpm0.577 & 2.361\tpm0.554 \\
\textbf{DCINN} & 35.591\tpm1.676 & 0.962\tpm0.009 & 0.975\tpm0.017 & 3.558\tpm0.691 & 2.669\tpm0.585 \\
\textbf{FAME} & 34.326\tpm1.743 & 0.948\tpm0.013 & 0.963\tpm0.037 & 4.250\tpm0.761 & 3.112\tpm0.595 \\
\textbf{SFINet++} & 35.636\tpm1.669 & 0.963\tpm0.009 & 0.976\tpm0.018 & 3.610\tpm0.663 & 2.627\tpm0.553 \\
\textbf{ViTPan} & 36.472\tpm1.824 & 0.967\tpm0.009 & 0.978\tpm0.017 & 3.187\tpm0.555 & 2.432\tpm0.525 \\
\textbf{HyperTransformer}& 36.135\tpm1.631 & 0.965\tpm0.009 & 0.976\tpm0.016 & 3.370\tpm0.496 & 2.497\tpm0.598 \\
\textbf{DCPNet} & 36.949\tpm1.736 & 0.968\tpm0.009 & 0.979\tpm0.016 & 3.104\tpm0.534 & 2.294\tpm0.490 \\
\textbf{MSCSCformer} & 36.901\tpm1.759 & 0.970\tpm0.009 & 0.980\tpm0.015 & 3.133\tpm0.547 & 2.319\tpm0.505 \\
\textbf{FSGformer} & \textit{37.964}\tpm\textit{1.774} & \textit{0.976}\tpm\textit{0.007} & \textit{0.985}\tpm\textit{0.013} & \textit{2.788}\tpm\textit{0.503} & \textit{2.035}\tpm\textit{0.440} \\
\midrule
\textbf{CGFformer}$($Ours$)$ & \textbf{38.348}\tpm\textbf{1.351} & \textbf{0.978}\tpm\textbf{0.005} & \textbf{0.986}\tpm\textbf{0.011} & \textbf{2.569}\tpm\textbf{0.452} & \textbf{1.869}\tpm\textbf{0.413} \\
\bottomrule
\end{tabular}
\end{table}

\begin{table}[ht]
\caption{Quantitative results of the fusion model on the GaoFen-2 Reduced-Resolution dataset. Bold shows the best result, and italicized values indicate the second-best method. }
\centering
\fontsize{9.2pt}{12pt}\selectfont
\setlength{\tabcolsep}{1.95pt}

\newcommand{\tpm}{$\mkern 1mu \pm \mkern 1mu$}

\begin{tabular}{@{} cccccc @{}}
\toprule
\textbf{Method} & \textbf{PSNR↑} & \textbf{SSIM↑} & \textbf{SCC↑} & \textbf{SAM↓} & \textbf{ERGAS↓} \\
\midrule
\textbf{GS}          & 31.526\tpm1.752 & 0.884\tpm0.033 & 0.930\tpm0.026 & 2.088\tpm0.382 & 2.420\tpm0.405 \\
\textbf{BDSD}        & 34.958\tpm2.011 & 0.903\tpm0.030 & 0.964\tpm0.018 & 1.725\tpm0.312 & 1.695\tpm0.390 \\
\textbf{PRACS}        & 35.222\tpm1.760 & 0.910\tpm0.023 & 0.964\tpm0.020 & 1.714\tpm0.312 & 1.646\tpm0.342 \\
\textbf{MTF-GLP}    & 34.544\tpm2.260 & 0.884\tpm0.039 & 0.960\tpm0.020 & 1.698\tpm0.353 & 1.722\tpm0.357 \\
\textbf{MF}          & 34.361\tpm1.801 & 0.897\tpm0.027 & 0.962\tpm0.015 & 1.685\tpm0.311 & 1.757\tpm0.421 \\
\textbf{PanNet}      & 38.880\tpm1.854 & 0.958\tpm0.010 & 0.985\tpm0.007 & 1.124\tpm0.198 & 1.771\tpm0.300 \\
\textbf{FusionNet}   & 38.946\tpm1.956 & 0.957\tpm0.011 & 0.985\tpm0.008 & 1.039\tpm0.200 & 1.425\tpm0.245 \\
\textbf{GPPNN}       & 36.336\tpm1.704 & 0.942\tpm0.014 & 0.973\tpm0.012 & 1.424\tpm0.225 & 1.083\tpm0.203 \\
\textbf{HLF-Net}     & 40.074\tpm1.605 & 0.968\tpm0.006 & 0.989\tpm0.005 & 1.014\tpm0.155 & 0.930\tpm0.148 \\
\textbf{MD$^3$Net}   & 43.750\tpm1.487 & 0.984\tpm0.004 & 0.995\tpm0.003 & 0.682\tpm0.116 & 0.601\tpm0.100 \\
\textbf{DCINN}       & 42.614\tpm1.462 & 0.980\tpm0.004 & 0.993\tpm0.003 & 0.767\tpm0.135 & 0.669\tpm0.109 \\
\textbf{FAME}        & 40.232\tpm1.501 & 0.971\tpm0.005 & 0.990\tpm0.005 & 0.875\tpm0.155 & 0.895\tpm0.147 \\
\textbf{SFINet++}    & 39.504\tpm1.667 & 0.966\tpm0.008 & 0.988\tpm0.005 & 1.001\tpm0.159 & 0.986\tpm0.152 \\
\textbf{ViTPan}      & 40.906\tpm1.529 & 0.973\tpm0.006 & 0.991\tpm0.004 & 0.866\tpm0.148 & 0.837\tpm0.134 \\
\textbf{HyperTransformer} & 41.776\tpm1.653 & 0.976\tpm0.005 & 0.992\tpm0.004 & 0.817\tpm0.146 & 0.763\tpm0.126 \\
\textbf{DCPNet}      & 42.725\tpm1.420 & 0.981\tpm0.004 & 0.994\tpm0.003 & 0.771\tpm0.099 & 0.686\tpm0.084 \\
\textbf{MSCSCformer} & 43.290\tpm1.424 & 0.982\tpm0.004 & 0.994\tpm0.003 & 0.731\tpm0.123 & 0.635\tpm0.095 \\
\textbf{FSGformer}   & \textit{45.050\tpm1.488} & \textit{0.987\tpm0.003} & \textit{0.996\tpm0.002} & \textit{0.597\tpm0.101} & \textit{0.520\tpm0.082} \\
\midrule
\textbf{CGFformer}$($Ours$)$     & \textbf{45.522\tpm1.328} & \textbf{0.989\tpm0.002} & \textbf{0.997\tpm0.001} & \textbf{0.575\tpm0.090} & \textbf{0.500\tpm0.075} \\
\bottomrule
\end{tabular}
\label{tab:tab3}
\end{table}

Our proposed model is compared with other existing methods. Specifically, traditional methods include GS \cite{ref56}, BDSD \cite{ref57}, PRACS \cite{ref8}, MTF-GLP \cite{ref59}, and MF \cite{ref60}; while deep learning-based methods include PanNet \cite{ref26}, FusionNet \cite{ref24}, GPPNN \cite{ref61}, HLF-Net \cite{ref31}, MD$^3$Net \cite{ref62}, DCINN \cite{ref27}, FAME \cite{ref35}, SFINet++ \cite{ref29}, ViTPan \cite{ref45}, HyperTransformer \cite{ref48}, DCPNet \cite{ref63}, MSCSCformer \cite{ref64}, and FSGformer \cite{ref36}. For each comparison model, we first adopt the parameter settings and adjustment strategies specified in the corresponding papers. To ensure fairness, all other settings for each method are set to their optimal values. All experiments are conducted on an NVIDIA RTX 3090 GPU.

For low-resolution fusion results, common quantitative evaluation metrics are used including peak signal-to-noise ratio (PSNR), structural similarity (SSIM), spatial correlation coefficient (SCC), spectral angle mapper (SAM), and Erreur Relative Globale Adimensionnelle de Synthèse (ERGAS) \cite{ref69}. For full-resolution fusion results, the hybrid quality with no reference (HQNR), spectral distortion index $D_\lambda$, and spatial distortion index $D_s$ are used \cite{ref71}.

\begin{figure}[t]
  \centering
  \includegraphics[width=1.0\textwidth]{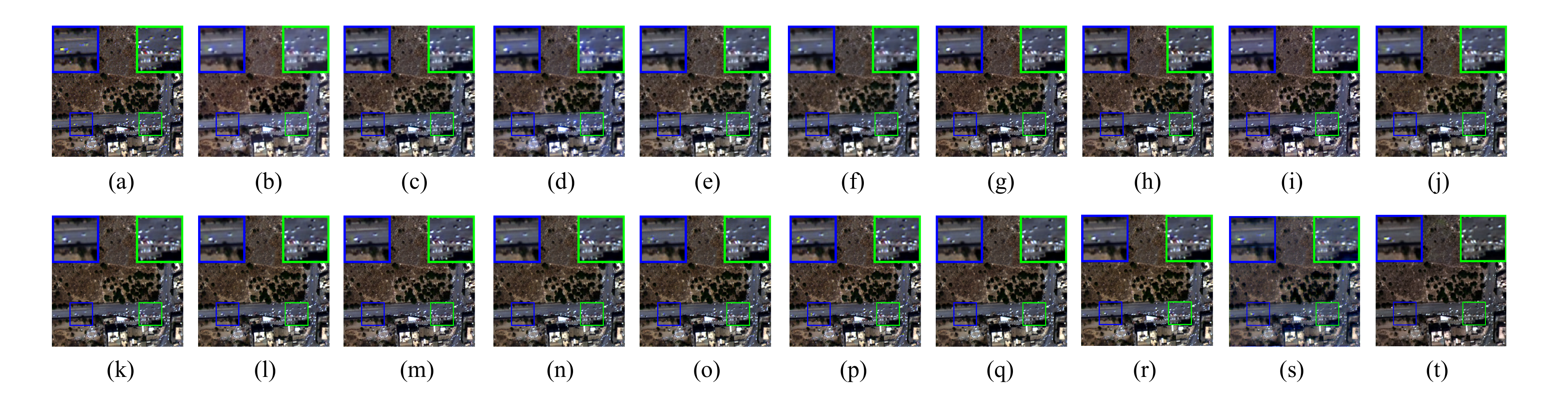}
  \caption{Visualization of different methods on the WorldView-3 reduced-resolution dataset. (a) GT. (b) GS. (c) BDSD. (d) PRACS. (e) MTF-GLP. (f) MF. (g) PanNet. (h) FusionNet. (i) GPPNN. (j) HLF-Net. (k) MD$^3$Net. (l) DCINN. (m) FAME. (n) SFINet++. (o) ViTPan. (p) HyperTransformer. (q) DCPNet. (r) MSCSCformer. (s) FSGformer. (t) CGFformer(Ours).}
  \label{fig:fig6}
\end{figure}

\begin{figure}[ht]
  \centering
  \includegraphics[width=1.0\textwidth]{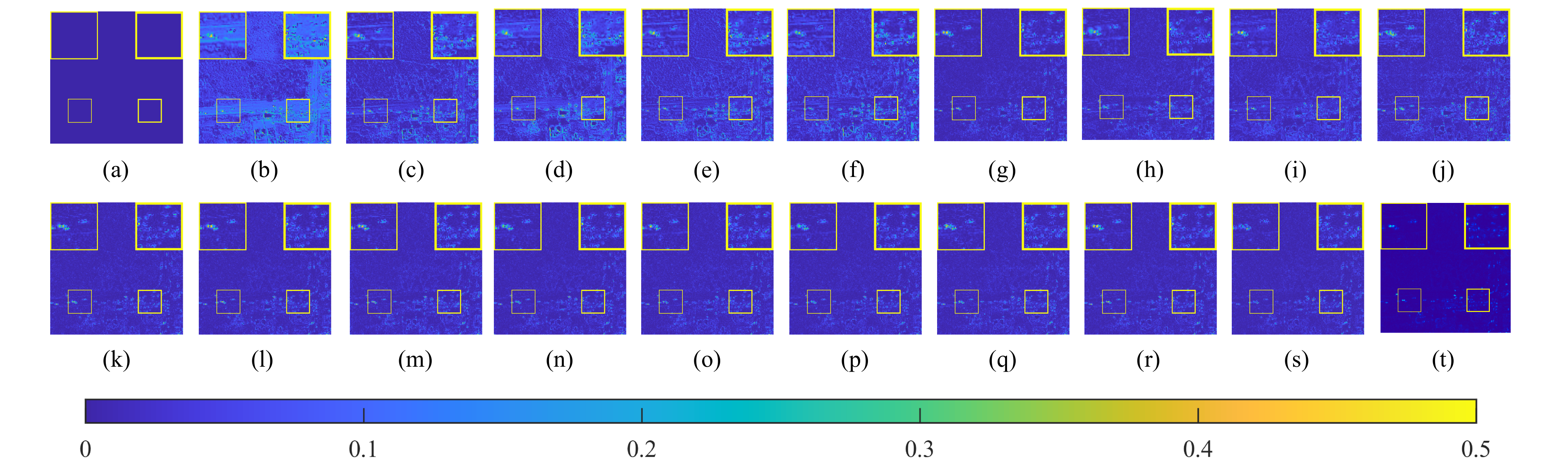}
  \caption{Mean absolute error maps between GT images and fused products on the WorldView-3 reduced-resolution dataset. (a) GT. (b) GS. (c) BDSD. (d) PRACS. (e) MTF-GLP. (f) MF. (g) PanNet. (h) FusionNet. (i) GPPNN. (j) HLF-Net. (k) MD$^3$Net. (l) DCINN. (m) FAME. (n) SFINet++. (o) ViTPan. (p) HyperTransformer. (q) DCPNet. (r) MSCSCformer. (s) FSGformer. (t) CGFformer(Ours).}
  \label{fig:fig7}
\end{figure}

\subsection{Reduced-Resolution Assessment}
\textit{1) Quantitative Analysis: }Table \ref{tab:tab2} presents the quantitative results of 19 methods on the WorldView-3 reduced-resolution dataset, with the best and second-best values in bold and italicized. Deep learning approaches generally outperform traditional methods, the proposed network achieves the best performance across all metrics. Specifically, it achieves PSNR gain of 0.383 dB over the second-best method and maintains SAM below 2.6, demonstrating superior spectral fidelity and spatial structure preservation. These consistent improvements stem from the proposed adaptive frequency separation, which enables spatially adaptive decomposition, and the mutual frequency-guidance denoising, which suppresses comprehensive noise to ensure high-quality spatial-frequency integration. Results on the GaoFen-2 dataset as shown in Table \ref{tab:tab2} confirm these advantages, further validating the robustness of the proposed frequency-based fusion framework.

\begin{figure}[t]
  \centering
  \includegraphics[width=1.0\textwidth]{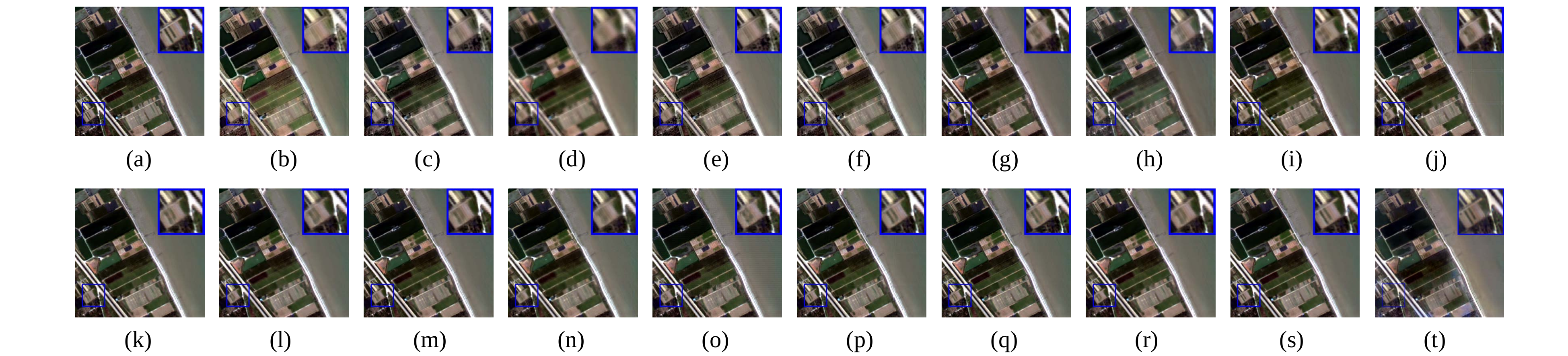}
  \caption{Visualization of different methods on the GaoFen-2 reduced-resolution dataset. (a) GT. (b) GS. (c) BDSD. (d) PRACS. (e) MTF-GLP. (f) MF. (g) PanNet. (h) FusionNet. (i) GPPNN. (j) HLF-Net. (k) MD$^3$Net. (l) DCINN. (m) FAME. (n) SFINet++. (o) ViTPan. (p) HyperTransformer. (q) DCPNet. (r) MSCSCformer. (s) FSGformer. (t) CGFformer(Ours).}
  \label{fig:fig8}
\end{figure}

\begin{figure}[t]
  \centering
  \includegraphics[width=1.0\textwidth]{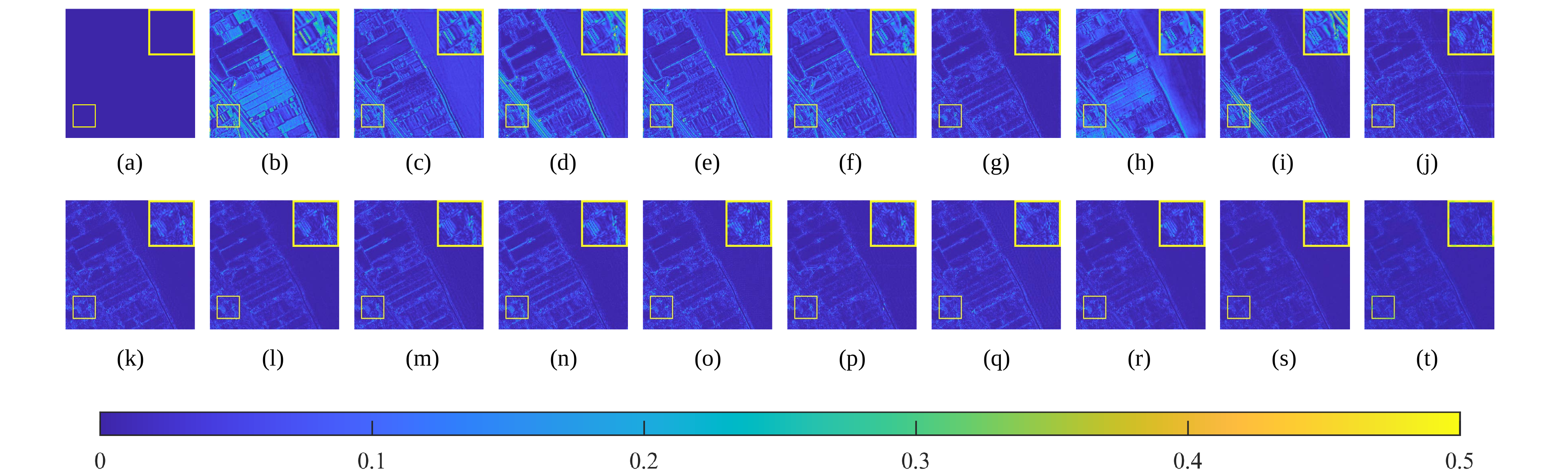}
  \caption{Mean absolute error maps between GT images and fused products on the GaoFen-2 reduced-resolution dataset. (a) GT. (b) GS. (c) BDSD. (d) PRACS. (e) MTF-GLP. (f) MF. (g) PanNet. (h) FusionNet. (i) GPPNN. (j) HLF-Net. (k) MD$^3$Net. (l) DCINN. (m) FAME. (n) SFINet++. (o) ViTPan. (p) HyperTransformer. (q) DCPNet. (r) MSCSCformer. (s) FSGformer. (t) CGFformer(Ours).}
  \label{fig:fig9}
\end{figure}

\textit{2) Qualitative Analysis: }Fig. \ref{fig:fig6} displays the visual comparisons on the WorldView-3 dataset. Compared to the blurred structures and degraded boundaries in traditional shown in Fig. \ref{fig:fig6}(b)–(f)) and most deep learning-based methods shown in Fig. \ref{fig:fig6}(g)–(s), the proposed approach shown in Fig. \ref{fig:fig6}(t) reconstructs sharper contours and finer details, closely matching the ground truth shown in Fig. \ref{fig:fig6}(a). The corresponding residual maps shown in Fig. \ref{fig:fig7} corroborate this; the residuals of the proposed method are closer to blue across the entire scene, indicating minimal reconstruction errors. This structural superiority is primarily attributed to the adaptive frequency separation mitigating inaccurate decomposition, and the bidirectional frequency-guidance denoising minimizing noise accumulation. Evaluations on the GaoFen-2 dataset shown in Figs. \ref{fig:fig8} and Figs. \ref{fig:fig9} demonstrate similar benefits. Amidst complex scenes, the proposed method preserves sharper texture boundaries and more accurate colors than competitors, which is further evidenced by the lower error intensities in the residual maps.

\begin{table}[t]
\caption{Quantitative results of the fusion model on WorldView-3 and GaoFen-2 Full-Resolution datasets. Bold shows the best result, and italicized values indicate the second-best method. }
\centering
\fontsize{9.2pt}{10pt}\selectfont
\begin{adjustbox}{center, max width=1.0\textwidth}
\setlength{\tabcolsep}{2.5pt}
\begin{tabular}{>{\centering\arraybackslash}c ccc ccc}
\toprule
\multirow{2}{*}{\textbf{Method}} & \multicolumn{3}{c}{\textbf{WorldView-3}} & \multicolumn{3}{c}{\textbf{GaoFen-2}} \\
\cmidrule(lr){2-4} \cmidrule(lr){5-7}
& \textbf{HQNR↑} & \textbf{$D_\lambda$↓} & \textbf{$D_s$↓} & \textbf{HQNR↑} & \textbf{$D_\lambda$↓} & \textbf{$D_s$↓} \\
\midrule
\textbf{GS} & 0.849 & 0.093 & 0.065 & 0.622 & 0.218 & 0.208 \\
\textbf{BDSD} & 0.870 & 0.062 & 0.073 & 0.781 & 0.076 & 0.155 \\
\textbf{PRACS} & 0.920 & 0.037 & 0.046 & 0.826 & 0.059 & 0.122 \\
\textbf{MTF-GLP} & 0.909 & 0.022 & 0.071 & 0.806 & 0.043 & 0.158 \\
\textbf{MF} & 0.912 & 0.028 & 0.063 & 0.799 & 0.054 & 0.156 \\
\textbf{PanNet} & 0.876 & 0.059 & 0.070 & 0.896 & 0.030 & 0.076 \\
\textbf{FusionNet} & 0.926 & 0.028 & 0.048 & 0.862 & 0.037 & 0.105 \\
\textbf{GPPNN} & 0.851 & 0.075 & 0.081 & 0.855 & 0.054 & 0.096 \\
\textbf{HLF-Net} & 0.905 & 0.046 & 0.052 & 0.897 & 0.032 & 0.072 \\
\textbf{MD$^3$Net} & 0.936 & 0.022 & 0.044 & 0.936 & 0.019 & 0.046 \\
\textbf{DCINN} & 0.918 & 0.027 & 0.057 & 0.934 & 0.019 & 0.048 \\
\textbf{FAME} & 0.916 & 0.031 & 0.055 & 0.930 & 0.024 & 0.049 \\
\textbf{SFINet++} & 0.919 & 0.029 & 0.054 & 0.898 & 0.028 & 0.075 \\
\textbf{ViTPan} & 0.915 & 0.038 & 0.050 & 0.911 & 0.035 & 0.056 \\
\textbf{HyperTransformer} & 0.946 & 0.022 & 0.033 & 0.927 & 0.029 & 0.045 \\
\textbf{DCPNet} & 0.920 & 0.041 & 0.041 & 0.917 & 0.026 & 0.059 \\
\textbf{MSCSCformer} & 0.934 & 0.022 & 0.046 & 0.948 & 0.023 & 0.030 \\
\textbf{FSGformer} & \textit{0.960} & \textit{0.016} & \textit{0.024} & \textit{0.958} & \textit{0.019} & \textit{0.024} \\
\midrule
\textbf{CGFformer}$($Ours$)$ & \textbf{0.961} & \textbf{0.014} & \textbf{0.022} & \textbf{0.961} & \textbf{0.017} & \textbf{0.023} \\
\bottomrule
\end{tabular}
\end{adjustbox}
\label{tab:tab4}
\end{table}

\subsection{Full-resolution Assessment}
\textit{1) Quantitative Analysis: }Table \ref{tab:tab4} presents the quantitative results on two full-resolution datasets, with the best and second-best values in bold and italicized. While deep learning approaches generally surpass traditional methods, the proposed network consistently achieves optimal performance across all evaluation metrics. Specifically, it achieves the best spectral quality index $D_\lambda$ and spatial quality index $D_s$ on both the GaoFen-2 and WorldView-3 datasets. These results demonstrate its superior capability in preserving spectral fidelity and enhancing spatial details in the absence of reference images.

\begin{figure}[t]
  \centering
  \includegraphics[width=1.0\textwidth]{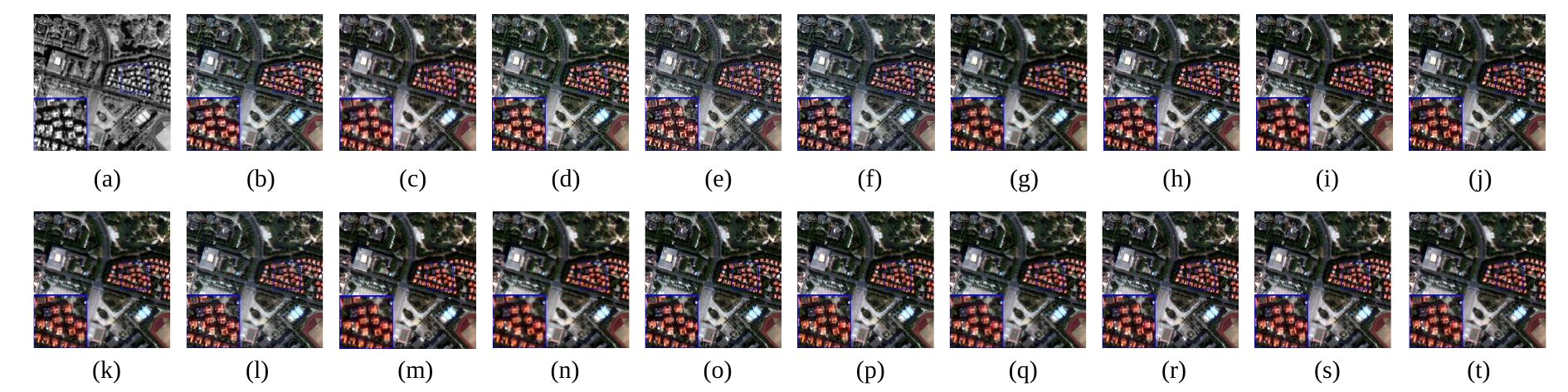}
  \caption{Visualization of different methods on the GaoFen-2 full-resolution dataset. (a) GT. (b) GS. (c) BDSD. (d) PRACS. (e) MTF-GLP. (f) MF. (g) PanNet. (h) FusionNet. (i) GPPNN. (j) HLF-Net. (k) MD$^3$Net. (l) DCINN. (m) FAME. (n) SFINet++. (o) ViTPan. (p) HyperTransformer. (q) DCPNet. (r) MSCSCformer. (s) FSGformer. (t) CGFformer(Ours).}
  \label{fig:fig10}
\end{figure}

\textit{2) Qualitative Analysis: }Fig. \ref{fig:fig10} illustrates the visual comparisons on the GaoFen-2 full-resolution dataset. As shown in the magnified regions, the proposed method synthesizes more natural appearances, resulting sharper structural details and higher spectral consistency than competing approaches. This visual superiority further validates the effectiveness of the proposed framework in high-quality spatial-frequency integration.

\subsection{Ablation Study}
\subsubsection{Frequency Separation Analysis}

\begin{figure}[ht]
  \centering
  \includegraphics[width=\textwidth]{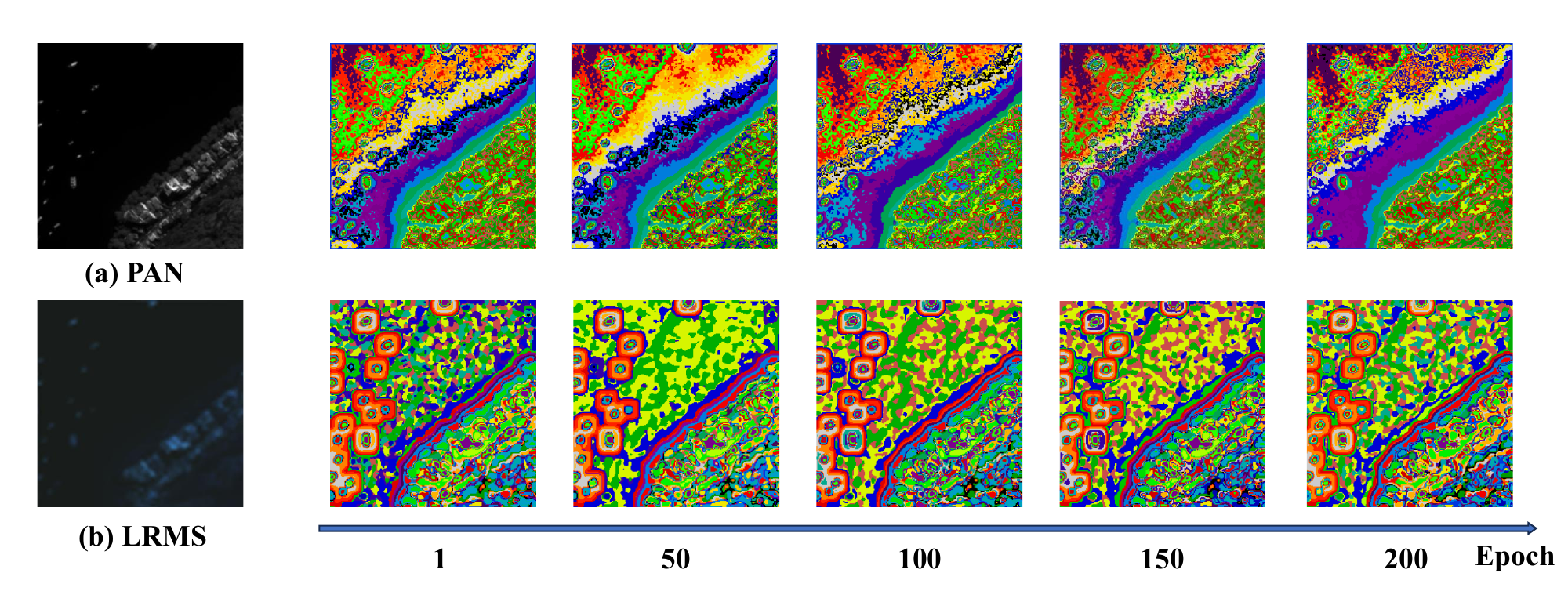}
  \caption{Visual representations of cluster index matrices in the model at different training epochs. (a) and (b) show the visual appearance of the raw PAN and LRMS input images. The right section displays the clustering results on these raw inputs without being transformed by convolution layers, where the upper row corresponds to the PAN image and the lower row corresponds to the LRMS image. The color indicates the cluster to which the pixel belongs.}
  \label{fig:fig11}
\end{figure}

Fig. \ref{fig:fig11} visualizes the clustering maps generated by the content-adaptive non-local (CAN) block to verify its spatial adaptability. Shallow-layer clusters primarily capture low-level color and texture similarities, whereas deep-layer clusters encode high-level semantic information. The convergence of the clustering index matrix during training indicates the progressive joint modeling of local details and non-local semantics. Furthermore, similar pixels naturally aggregate into large clusters, outliers are isolated into smaller subsets, and boundary regions are clearly demarcated through the zero-padding mechanism during unfolding.

Table \ref{tab:tab5} presents quantitative comparisons when replacing the CAN block with classical low-pass filters includes Gaussian, Fourier, and local-neighborhood \cite{ref36} patterns, keeping subsequent steps unchanged. The proposed CAFS module consistently outperforms these alternatives across all metrics. This confirms that the CAN-based filter better aligns with the inherent frequency distribution by exploiting both local structures and non-local similarities, underscoring the necessity of spatial adaptivity in frequency separation.

\begin{figure}[ht]
  \centering
  \includegraphics[width=0.6\textwidth]{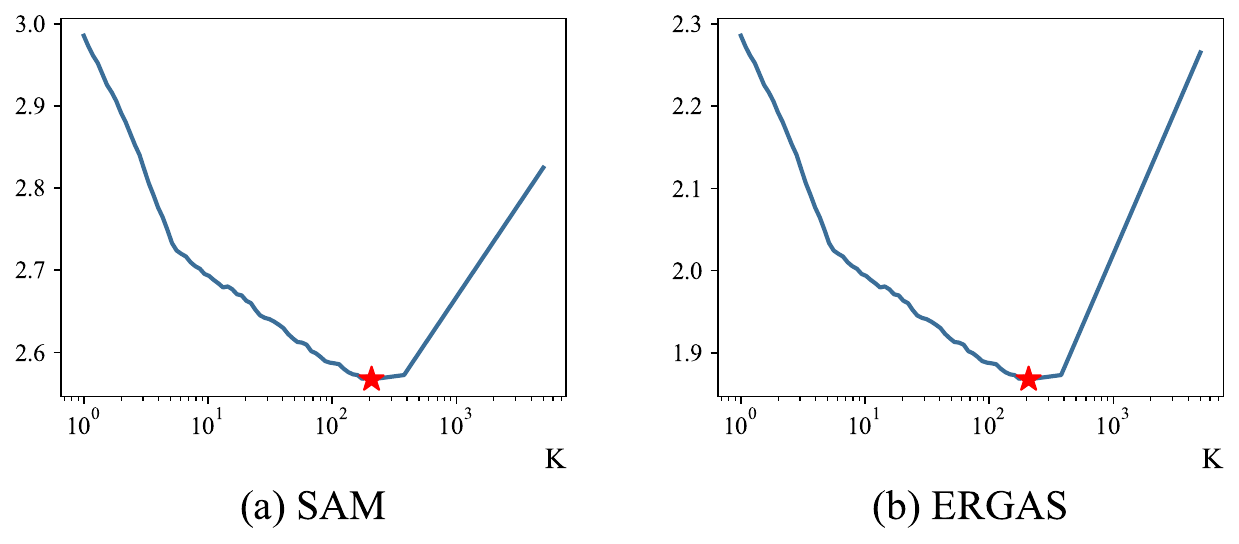}
  \caption{Variations of SAM, ERGAS on the WV3 reduced-resolution dataset with changing cluster number $K$. The optimal metrics are obtained around $K = 128$.}
  \label{fig:fig12}
\end{figure}

We further ablate the cluster number $K$ in the K-means algorithm. The network is trained with $K=32$ and evaluated using varying $K$ values during inference. As shown in Fig. \ref{fig:fig12}, performance initially improves as $K$ increases but degrades when $K$ becomes excessive. A small $K$ groups dissimilar pixels, approximating global filtering and hindering adaptive kernel generation. Conversely, an overly large $K$ results sparse clusters, impeding robust non-local aggregation. The curve extremes effectively represent the limiting cases of global convolution and fully spatially adaptive kernels, confirming that appropriate clustering is crucial for non-local information transmission. Notably, the optimal test $K$ is approximately four times the training $K$. This discrepancy arises because test images encompass larger spatial extents and more diverse frequency distributions than the training patches.

\begin{table}[t]
\caption{Quantitative results of frequency separation ablation experiments on WorldView-3 dataset. The best results are highlighted in bold.}
\centering
\fontsize{9.2pt}{10pt}\selectfont
\begin{adjustbox}{center, max width=1.0\textwidth}
\begin{tabular}{>{\centering\arraybackslash}cccccc}
\toprule
\textbf{Frequency} & \textbf{PSNR↑} & \textbf{SSIM↑} & \textbf{SCC↑} & \textbf{SAM↓} & \textbf{ERGAS↓} \\
\midrule
Local adaptation & 38.144 & 0.977 & 0.981 & 2.599 & 1.990 \\
Fourier transform & 37.519 & 0.969 & 0.977 & 2.893 & 2.262 \\
Gaussian blur & 37.216 & 0.972 & 0.979 & 2.924 & 2.326 \\
\textbf{Ours} & \textbf{38.348} & \textbf{0.978} & \textbf{0.986} & \textbf{2.569} & \textbf{1.869} \\
\bottomrule
\end{tabular}
\end{adjustbox}
\label{tab:tab5}
\end{table}

\begin{table}[t]
\caption{Quantitative results of ablation experiments of guided fusion mechanism on WorldView-3 dataset. The best results are highlighted in bold.}
\centering
\fontsize{9.2pt}{10pt}\selectfont
\begin{adjustbox}{center, max width=1.0\textwidth}
\begin{tabular}{>{\centering\arraybackslash}cccccc}
\toprule
\textbf{Pattern} & \textbf{PSNR↑} & \textbf{SSIM↑} & \textbf{SCC↑} & \textbf{SAM↓} & \textbf{ERGAS↓} \\
\midrule
Without Denoising & 37.959 & 0.974 & 0.974 & 2.809 & 2.214 \\
Without Frequency & 37.234 & 0.972 & 0.969 & 2.798 & 2.419 \\
Without Guidance & 36.919 & 0.969 & 0.970 & 3.020 & 2.601 \\
Without Spatial & 37.919 & 0.976 & 0.980 & 2.610 & 1.933 \\
\textbf{Ours} & \textbf{38.348} & \textbf{0.978} & \textbf{0.986} & \textbf{2.569} & \textbf{1.869} \\
\bottomrule
\end{tabular}
\end{adjustbox}
\label{tab:tab6}
\end{table}

\begin{table}[t]
\caption{Quantitative results of network module ablation experiments on WorldView-3 dataset. The best results are highlighted in bold.}
\centering
\fontsize{9.2pt}{10pt}\selectfont
\begin{adjustbox}{center, max width=1.0\textwidth}
\begin{tabular}{>{\centering\arraybackslash}cccccc}
\toprule
\textbf{Pattern} & \textbf{PSNR↑} & \textbf{SSIM↑} & \textbf{SCC↑} & \textbf{SAM↓} & \textbf{ERGAS↓} \\
\midrule
Without DSR & 37.010 & 0.971 & 0.971 & 3.188 & 2.322 \\
Without SFA-S & 37.921 & 0.972 & 0.983 & 2.880 & 2.010 \\
Without SFA-F & 36.710 & 0.969 & 0.978 & 2.925 & 2.101 \\
Without SFA & 35.981 & 0.969 & 0.975 & 3.109 & 2.282 \\
\textbf{Ours} & \textbf{38.348} & \textbf{0.978} & \textbf{0.986} & \textbf{2.569} & \textbf{1.869} \\
\bottomrule
\end{tabular}
\end{adjustbox}
\label{tab:tab7}
\end{table}

\subsubsection{Denoising and Fusion Mechanisms Analysis}
Table \ref{tab:tab6} presents the ablation study on the WorldView-3 dataset, evaluating the individual removal of the denoising calibration, mutual frequency guidance in DSR module, and spatial-frequency fusion in SFA module. Each omission yields clear performance degradation. Specifically, removing the mutual guidance weakens noise suppression, validating the DSR module's reliance on cross-frequency interaction to refine frequency components. Eliminating the spatial-frequency fusion compromises HRMS reconstruction, underscoring the role of SFA module in the complementary integration of spatial structures and frequency details. Furthermore, excluding the NCB calibration impairs the filtering of redundant responses, confirming its necessity for stable frequency representations.

\subsubsection{Network Module Analysis}
Table \ref{tab:tab7} details the module-level ablations on the WorldView-3 dataset. To ablate the SFA module, high- and low-frequency outputs from the DSR module are directly summed and convolved to maintain the inference pipeline. Each module's omission yields a clear performance decline, highlighting their distinct structural roles. Specifically, removing the DSR module disrupts high- and low-frequency interactions, compromising cross-frequency consistency. Excluding the SFA-S branch degrades spatial texture representations, while ablating the SFA-F branch hinders multi-directional feature aggregation. Consequently, omitting the entire SFA module severely impairs the spatial-frequency fusion quality, validating the necessity and complementary synergy of all proposed components.

\section{Conclusion}
In this article, we present a Transformer-based pansharpening framework that generates high-quality HRMS images by exploiting the varying characteristics and interactions of spatial and frequency components. Our approach fully leverages the advantages of adaptive frequency separation and spatial-frequency fusion, offering an effective solution for remote sensing pansharpening. Specifically, the CAFS module models both local neighborhoods and non-local similarity to achieve spatially adaptive decomposition of high- and low-frequency components. The DSR module integrates frequency correlations with attention mechanisms, enabling accurate suppression of frequency-relevant and frequency-irrelevant noise. Moreover, the SFA-S branch strengthen structural detail and the SFA-F branch promote comprehensive interaction between spatial and frequency information, facilitating efficient feature learning. Extensive experiments on multiple datasets confirm the effectiveness and robustness of the proposed method and future work will extend this framework to cross-modal fusion tasks.

\section*{Data Availability}
The GaoFen-2 and WorldView-3 datasets analyzed in this study are publicly available at https://github.com/liangjiandeng/PanCollection. 
The code of CGFformer is available at https://github.com/PatrickNod/CGFformer.

\section*{Acknowledgements}
This work was supported in part by the National Natural Science Foundation of China under Grant 12201490, in part by the China Postdoctoral Science Foundation under Grant 2025M773093, and in part by the Postdoctoral Fellowship Program of CPSF under Grant GZC20252028.

\section*{CRediT authorship contribution statement}
\textbf{Zijian Zhou:} Conceptualization, Methodology, Software, Writing - original draft. \textbf{Jianing Zhang:} Data curation, Validation, Investigation, Software. \textbf{Kai Sun:} Supervision, Funding acquisition, Writing - review \& editing. \textbf{Xiangyu Zhao:} Formal analysis, Validation. \textbf{Chunxia Zhang:} Validation, Writing - review \& editing. \textbf{Xiangyong Cao:} Visualization, Writing - review \& editing.

\section*{Declaration of competing interest}
The authors declare that they have no known competing financial interests or personal relationships that could have appeared to influence the work reported in this paper.


\begin{thebibliography}{00}

\bibitem{ref2}
D. Tuia, J. Munoz-Mari, G. Camps-Valls, Remote sensing image segmentation by active queries, Pattern Recognit. 45 (2012) 2180--2192.

\bibitem{ref4}
A. Troya-Galvis, P. Gançarski, L. Berti-Équille, Remote sensing image analysis by aggregation of segmentation-classification collaborative agents, Pattern Recognit. 73 (2018) 259--274.

\bibitem{ref5}
D. Wang, P. Qiu, B. Wan, Z. Cao, Q. Zhang, Mapping $\alpha$- and $\beta$-diversity of mangrove forests with multispectral and hyperspectral images, Remote Sens. Environ. 275 (2022) 113021.

\bibitem{ref6}
L.-J. Deng, et al., Machine Learning in Pansharpening: A benchmark, from shallow to deep networks, IEEE Geosci. Remote Sens. Mag. 10 (3) (2022) 279--315.

\bibitem{ref7}
X. Meng, H. Shen, H. Li, L. Zhang, R. Fu, Review of the pansharpening methods for remote sensing images based on the idea of meta-analysis: Practical discussion and challenges, Inf. Fusion 46 (2019) 102--113.

\bibitem{ref8}
J. Choi, K. Yu, Y. Kim, A New Adaptive Component-Substitution-Based Satellite Image Fusion by Using Partial Replacement, IEEE Trans. Geosci. Remote Sens. 49 (1) (2011) 295--309.

\bibitem{ref11}
G. Vivone, R. Restaino, M. Dalla Mura, G. Licciardi, J. Chanussot, Contrast and Error-Based Fusion Schemes for Multispectral Image Pansharpening, IEEE Geosci. Remote Sens. Lett. 11 (5) (2014) 930--934.

\bibitem{ref14}
X. Fu, Z. Lin, Y. Huang, X. Ding, A Variational Pan-Sharpening With Local Gradient Constraints, CVPR (2019) 10257--10266.

\bibitem{ref17}
G. Vivone, et al., A Critical Comparison Among Pansharpening Algorithms, IEEE Trans. Geosci. Remote Sens. 53 (5) (2015) 2565--2586.

\bibitem{ref18}
K. Amolins, Y. Zhang, P. Dare, Wavelet based image fusion techniques - An introduction, review and comparison, ISPRS J. Photogramm. Remote Sens. 62 (4) (2007) 249--263.

\bibitem{ref19}
C. S. Yilmaz, V. Yilmaz, O. Gungor, A theoretical and practical survey of image fusion methods for multispectral pansharpening, Inf. Fusion 79 (2022) 1--43.

\bibitem{ref20}
G. Masi, D. Cozzolino, L. Verdoliva, G. Scarpa, Pansharpening by Convolutional Neural Networks, Remote Sens. 8 (7) (2016) 594.

\bibitem{ref21}
J. Li, K. Zheng, J. Yao, L. Gao, D. Hong, Deep Unsupervised Blind Hyperspectral and Multispectral Data Fusion, IEEE Geosci. Remote Sens. Lett. 19 (2022) 1--5.

\bibitem{ref22}
Z.-R. Jin, T.-J. Zhang, T.-X. Jiang, G. Vivone, L.-J. Deng, LAGConv: Local-Context Adaptive Convolution Kernels, AAAI (2022) 1113--1121.

\bibitem{ref23}
L. He, et al., Pansharpening via Detail Injection Based Convolutional Neural Networks, IEEE J. Sel. Top. Appl. Earth Obs. Remote Sens. 12 (4) (2019) 1188--1204.

\bibitem{ref24}
L.-J. Deng, G. Vivone, C. Jin, J. Chanussot, Detail injection-based deep convolutional neural networks for pansharpening, IEEE Trans. Geosci. Remote Sens. 59 (2020) 6995--7010.

\bibitem{ref25}
H. Lu, Y. Yang, S. Huang, X. Chen, H. Su, W. Tu, Intensity mixture and band-adaptive detail fusion for pansharpening, Pattern Recognit. 139 (2023) 109434.

\bibitem{ref26}
J. Yang, X. Fu, Y. Hu, Y. Huang, X. Ding, J. Paisley, PanNet: A Deep Network Architecture for Pan-Sharpening, ICCV (2017) 5449--5457.

\bibitem{ref27}
W. Wang, L.-J. Deng, R. Ran, G. Vivone, A General Paradigm with Detail-Preserving Conditional Invertible Network for Image Fusion, Int. J. Comput. Vis. 132 (4) (2023) 1029--1054.

\bibitem{ref28}
M. Zhou, et al., Spatial-Frequency Domain Information Integration for Pan-Sharpening, ECCV (2022) 274--291.

\bibitem{ref29}
M. Zhou, et al., A General Spatial-Frequency Learning Framework for Multimodal Image Fusion, IEEE Trans. Pattern Anal. Mach. Intell. 47 (2025) 5281--5298.

\bibitem{ref30}
M. Fritsche, S. Gu, R. Timofte, Frequency Separation for Real-World Super-Resolution, ICCVW (2019) 3599--3608.

\bibitem{ref31}
W. Diao, F. Zhang, H. Wang, W. Wan, J. Sun, K. Zhang, HLF-Net: Pansharpening Based on High- and Low-Frequency Fusion Networks, IEEE Geosci. Remote Sens. Lett. 19 (2022) 1--5.

\bibitem{ref32}
X. Zou, F. Xiao, Z. Yu, et al., Delving Deeper into Anti-Aliasing in ConvNets, Int. J. Comput. Vis. 131 (2023) 67--81.

\bibitem{ref33}
M. Zhou, et al., Adaptively Learning Low-high Frequency Information Integration for Pan-sharpening, ACM Multimedia (2022) 3375--3384.

\bibitem{ref34}
Y. Xing, Y. Zhang, H. He, X. Zhang, Y. Zhang, Pansharpening via Frequency-Aware Fusion Network With Explicit Similarity Constraints, IEEE Trans. Geosci. Remote Sens. 61 (2023) 1--14.

\bibitem{ref35}
X. He, K. Yan, R. Li, C. Xie, J. Zhang, M. Zhou, Frequency-Adaptive Pan-Sharpening with Mixture of Experts, AAAI (2024) 2121--2129.

\bibitem{ref36}
Q. Liu, X. Zhao, Y. Qin, L. Li, J. Liu, FSGformer: Frequency Separation and Guidance Transformer for Pansharpening, IEEE Trans. Geosci. Remote Sens. 63 (2025) 1--16.

\bibitem{ref37}
Y. Duan, X. Wu, H. Deng, L.-J. Deng, Content-Adaptive Non-Local Convolution for Remote Sensing Pansharpening, CVPR (2024) 27738--27747.

\bibitem{ref38}
J. W. Gibbs, Fourier's series, Nature 59 (1539) (1899) 606.

\bibitem{ref100}
H. Mo, J. Jiang, Q. Wang, D. Yin, P. Dong, J. Tian, Frequency Attention Network: Blind Noise Removal for Real Images, ACCV (2020) 168--184.

\bibitem{ref39}
W. G. C. Bandara, J. M. J. Valanarasu, V. M. Patel, Hyperspectral Pansharpening Based on Improved Deep Image Prior and Residual Reconstruction, IEEE Trans. Geosci. Remote Sens. 60 (2022) 1--16.

\bibitem{ref40}
H. Lu, Y. Yang, S. Huang, R. Liu, H. Guo, MSAN: Multiscale self-attention network for pansharpening, Pattern Recognit. 162 (2025) 111441.

\bibitem{ref41}
Y. Yang, G. Yuan, J. Li, SFFNet: A Wavelet-Based Spatial and Frequency Domain Fusion Network for Remote Sensing Segmentation, IEEE Trans. Geosci. Remote Sens. 62 (2024) 1--17.

\bibitem{ref42}
K. S. Charan, G. Rochan Ravi, T. N. Shashank, C. Gururaj, Image Super-Resolution Using Convolutional Neural Network, MysuruCon (2022) 1--7.

\bibitem{ref43}
H. Zhang, H. Wang, X. Tian, J. Ma, P2Sharpen: A progressive pansharpening network, Inf. Fusion 91 (2023) 103--122.

\bibitem{ref44}
Q. Cao, L.-J. Deng, W. Wang, J. Hou, G. Vivone, Zero-shot semi-supervised learning for pansharpening, Inf. Fusion 101 (2024) 102001.

\bibitem{ref45}
X. Meng, N. Wang, F. Shao, S. Li, Vision Transformer for Pansharpening, IEEE Trans. Geosci. Remote Sens. 60 (2022) 1--14.

\bibitem{ref46}
H. Zhou, Q. Liu, Y. Wang, PanFormer: A Transformer Based Model for Pan-Sharpening, ICME (2022) 1--6.

\bibitem{ref47}
Z. Liu, et al., Swin Transformer: Hierarchical Vision Transformer using Shifted Windows, ICCV (2021) 10012--10022.

\bibitem{ref48}
W. G. C. Bandara, V. M. Patel, HyperTransformer: A textural and spectral feature fusion transformer for pansharpening, CVPR (2022) 1757--1767.

\bibitem{ref49}
S. W. Zamir, A. Arora, S. Khan, M. Hayat, F. S. Khan, M.-H. Yang, Restormer: Efficient Transformer for High-Resolution Image Restoration, CVPR (2022) 5728--5739.

\bibitem{ref53}
J. Huang, R. Huang, J. Xu, S. Peng, Y. Duan, L.J. Deng, Wavelet-assisted multi-frequency attention network for pansharpening, AAAI (2025) 3662--3670.

\bibitem{ref54}
Y. Ding, Y. Zhao, X. Shen, et al., Yinyang k-means: A drop-in replacement of the classic k-means with consistent speedup, ICML (2015) 579--587.

\bibitem{ref56}
C. A. Laben, B. V. Brower, Process for enhancing the spatial resolution of multispectral imagery using pan-sharpening, U.S. Patent 6,011,875 (2000).

\bibitem{ref57}
A. Garzelli, F. Nencini, L. Capobianco, Optimal MMSE Pan Sharpening of Very High Resolution Multispectral Images, IEEE Trans. Geosci. Remote Sens. 46 (1) (2008) 228--236.

\bibitem{ref59}
B. Aiazzi, L. Alparone, S. Baronti, A. Garzelli, M. Selva, MTF-tailored Multiscale Fusion of High-resolution MS and Pan Imagery, Photogramm. Eng. Remote Sens. 72 (5) (2006) 591--596.

\bibitem{ref60}
R. Restaino, G. Vivone, M. Dalla Mura, J. Chanussot, Fusion of Multispectral and Panchromatic Images Based on Morphological Operators, IEEE Trans. Image Process. 25 (6) (2016) 2882--2895.

\bibitem{ref61}
S. Xu, J. Zhang, Z. Zhao, K. Sun, J. Liu, C. Zhang, Deep Gradient Projection Networks for Pan-sharpening, CVPR (2021) 1366--1375.

\bibitem{ref62}
Y. Yan, J. Liu, S. Xu, Y. Wang, X. Cao, MD$^3$Net: Integrating Model-Driven and Data-Driven Approaches for Pansharpening, IEEE Trans. Geosci. Remote Sens. 60 (2022) 1--16.

\bibitem{ref63}
Y. Zhang, X. Yang, H. Li, M. Xie, Z. Yu, DCPNet: A Dual-Task Collaborative Promotion Network for Pansharpening, IEEE Trans. Geosci. Remote Sens. 62 (2024) 1--16.

\bibitem{ref64}
Y. Ye, T. Wang, F. Fang, G. Zhang, MSCSCformer: Multiscale Convolutional Sparse Coding-Based Transformer for Pansharpening, IEEE Trans. Geosci. Remote Sens. 62 (2024) 1--12.

\bibitem{ref69}
K. Zhang, F. Zhang, W. Wan, H. Yu, J. Sun, J. Del Ser, E. Elyan, A. Hussain, Panchromatic and multispectral image fusion for remote sensing and earth observation: Concepts, taxonomy, literature review, evaluation methodologies and challenges ahead, Inf. Fusion 93 (2023) 227--242.

\bibitem{ref71}
A. Arienzo, G. Vivone, A. Garzelli, L. Alparone, J. Chanussot, Full-resolution quality assessment of pansharpening: Theoretical and hands-on approaches, IEEE Geosci. Remote Sens. Mag. 10 (3) (2022) 168--201.

\end{thebibliography}
\end{document}